\documentclass[journal]{IEEEtran}
\usepackage{cite}

\usepackage[pdftex]{graphicx}
\usepackage{tikz,pgfplots}
\usetikzlibrary{positioning,shadows,arrows,trees,shapes,fit}
\usetikzlibrary{arrows.meta}
\pgfplotsset{compat=1.14}
\tikzstyle{vertex} = [fill,shape=circle,node distance=30pt]
\tikzstyle{edge} = [fill,opacity=.6,fill opacity=.5,line cap=round, line join=round, line width=10pt]
\tikzstyle{elabel} =  [fill,shape=circle,node distance=30pt,fill opacity=.9]
\definecolor{mygray}{gray}{0.95}
\pgfdeclarelayer{background}
\pgfsetlayers{background,main}
\usepackage[most]{tcolorbox}
\definecolor{aureolin}{rgb}{0.99, 0.93, 0.0}
\definecolor{asparagus}{rgb}{0.53, 0.66, 0.42}
\definecolor{capri}{rgb}{0.0, 0.75, 1.0}
\definecolor{darkpink}{rgb}{0.91, 0.33, 0.5}
\definecolor{eggplant}{rgb}{0.38, 0.25, 0.32}
\definecolor{fuchsia}{rgb}{1.0, 0.0, 1.0}
\definecolor{goldenrod}{rgb}{0.85, 0.65, 0.13}
\definecolor{green(pigment)}{rgb}{0.0, 0.65, 0.31}
\definecolor{salmon}{rgb}{1.0, 0.55, 0.41}
\definecolor{amber}{rgb}{1.0, 0.75, 0.0}
\definecolor{amethyst}{rgb}{0.6, 0.4, 0.8}
\definecolor{applegreen}{rgb}{0.55, 0.71, 0.0}
\definecolor{ashgrey}{rgb}{0.7, 0.75, 0.71}
\definecolor{babyblue}{rgb}{0.54, 0.81, 0.94}
\definecolor{bittersweet}{rgb}{1.0, 0.44, 0.37}
\definecolor{blue-green}{rgb}{0.0, 0.87, 0.87}
\definecolor{darkcerulean}{rgb}{0.03, 0.27, 0.49}
\definecolor{brilliantrose}{rgb}{1.0, 0.33, 0.64}
\definecolor{emerald}{rgb}{0.31, 0.78, 0.47}
\definecolor{sandstorm}{rgb}{0.93, 0.84, 0.25}
\definecolor{teal}{rgb}{0.0, 0.5, 0.5}
\definecolor{sealbrown}{rgb}{0.2, 0.08, 0.08}
\definecolor{bleudefrance}{rgb}{0.19, 0.55, 0.91}
\newcommand\redpt[1]{\fill[black] (#1) circle[radius=1mm];}

\ifCLASSINFOpdf
\else
\fi
\usepackage{amsmath}
\usepackage{amssymb}
\usepackage{mathrsfs}
\usepackage{mathtools}
\usepackage[utf8]{inputenc}
\usepackage[english]{babel}
\usepackage{amsthm}

\usepackage{algorithmic}

\usepackage[ruled]{algorithm}

\usepackage{array}
\newcommand{\R}{\mathbb{R}}

\newcommand{\g}{\mathcal{G}}
\newcommand{\gr}{G}
\newcommand{\Vs}{V}
\newcommand{\Es}{E}
\newcommand{\vx}{\mathbf{x}}
\newcommand{\mA}{\mathbf{A}}
\newcommand{\mL}{\mathbf{L}}
\newcommand{\mB}{\mathbf{B}}
\newcommand{\mZ}{\mathbf{0}}
\newcommand{\mcT}{\mathcal{T}}
\newcommand{\tmA}{\tilde{\mathbf{A}}}
\newcommand{\tmL}{\tilde{\mathbf{L}}}
\newcommand{\tg}{\tilde{\mathcal{G}}}
\newcommand{\tgr}{\tilde{G}}

\newcommand{\lam}{\lambda}
\newcommand{\tlam}{\tilde{\lambda}}
\newcommand{\dm}{D}
\newcommand{\dmh}{\mathcal{D}}
\newcommand{\Nnodes}{n}
\newcommand{\mD}{\mathbf{D}}
\newcommand{\mW}{\mathbf{W}}
\newcommand{\mH}{\mathbf{H}}
\newcommand{\mI}{\mathbf{I}}
\newcommand{\tA}{\textsf{A}}
\newcommand{\ttA}{\tilde{\textsf{A}}}
\newcommand{\tD}{\textsf{D}}
\newcommand{\tL}{\textsf{L}}
\newcommand{\ttL}{\tilde{\textsf{L}}}
\newcommand{\mLam}{\mathbf{\Lambda}}
\newcommand{\mV}{\mathbf{V}}
\newcommand{\evec}{\mathbf{v}}
\newcommand{\Prob}{\mathcal{P}}
\newcommand{\Gs}{\mathbb{G}}

\begin{document}
%
\title{Hypergraph Dissimilarity Measures}
%
%
%

\author{
        Amit~Surana,
        Can~Chen
        and~Indika~Rajapakse

\thanks{A. Surana is with Raytheon Technologies Research Center, East Hartford, CT 06108 USA (e-mail: amit.surana@rtx.com).}
\thanks{C. Chen is with the Department of Mathematics, University of Michigan, Ann Arbor, MI 48109 USA (e-mail: canc@umich.edu).}%
\thanks{I. Rajapakse is with the Department of Computational Medicine \& Bioinformatics, Medical School and the Department of Mathematics, University of Michigan, Ann Arbor, MI 48109 USA (e-mail: indikar@umich.edu).}
}

\maketitle

\begin{abstract}
 In this paper, we propose two novel approaches for hypergraph comparison. The first approach transforms the hypergraph into a graph representation for use of standard graph dissimilarity measures. The second approach exploits the mathematics of tensors to intrinsically capture multi-way relations. For each approach, we present measures that assess hypergraph dissimilarity at a specific scale or provide a more holistic multi-scale comparison. We test these measures on synthetic hypergraphs and apply them to biological datasets.

\end{abstract}


%
\IEEEpeerreviewmaketitle

\section{Introduction}
\IEEEPARstart{C}{omplex} systems in sociology, biology, cyber-security, telecommunications, and physical infrastructure are often represented as a set of entities, i.e. “vertices” with binary relationships or “edges," and hence are analyzed via graph theoretic methods. Graph models, while simple and to some degree universal, are limited to representing pairwise relationships between entities. However, real-world phenomena can be rich in multi-way relationships, dependencies between more than two variables, or properties of collections of more than two objects. Examples include computer networks where the dynamic relations are defined by packets exchanged over time between computers, co-authorship networks where relations are articles written by two or more authors, historical documents where multiple persons can be mentioned together, brain activity where multiple regions can be highly active at the same time, film actor networks, and protein-protein interaction networks, \cite{newman2018networks,battiston2020networks,benson2021higher,sweeney2021network}.


A hypergraph is a generalization of a graph in which its hyperedges can join any number of vertices \cite{berge1984hypergraphs}. Thus, hypergraphs can capture multi-way relationships unambiguously \cite{wolf2016advantages}, and are the natural representation of a broad range of systems mentioned above.  Although an expanding body of research attests to the increased utility of hypergraph-based analyses, many network science methods have been historically developed explicitly (and often, exclusively) for graph-based analyses and do not directly translate to hypergraphs. Consequently, new framework are being developed for representation, learning and analysis of hypergraphs, see \cite{battiston2020networks,benson2021higher} for a recent survey. These include techniques for converting hypergraphs into graphs and defining hypergraph Laplacian \cite{agarwal2006higher}, higher-order random walks-based hypergraph analysis \cite{aksoy2020hypernetwork}, and defining dynamics on hypergraphs \cite{carletti2020dynamical}.

As tensors \cite{Kolda06multilinearoperators} provide a natural framework to represent multi-dimensional patterns and capture higher-order interactions, they are finding increasing role in context of hypergraphs. For example, the spectral theory of graphs has been extended to hypergraphs using tensor eigenvalues \cite{banerjee2017spectra}, and authors in \cite{9119161} define notion of tensor entropy for uniform hypergraphs generalizing von Neumann entropy of a graph to hypergraphs. The problem of controllability of dynamics on hypergraphs is studied via tensor-based representation and nonlinear control theory in \cite{chen2021controllability}. Similar to above mentioned work, the goal of this paper is to extend the graph comparison framework to hypergraphs.

Comparison of structures such as modular communities, hubs, and trees yield insight into the generative mechanisms and functional properties of the graph. Graph comparison can be used for comparing brain or metabolic networks for different subjects, or the same subject before and after a treatment, and for characterizing the temporal network evolution during treatment\cite{donnat2018tracking}. Classification of graphs, for example in context of protein-protein interaction networks and online social networks can be facilitated via use of graph comparison measures \cite{yanardag2015deep}. Combined with a clustering algorithm, a graph comparison measure can be used to aggregate networks in a meaningful way and reveal redundancy in the data/networks \cite{de2015structural}. Graph comparison can be used for evaluating the accuracy of statistical or generative network models \cite{hunter2008goodness}, and can be further utilized as an objective function to drive the optimization procedure to fit graph models to data.

In order to compare graphs, a variety of  dissimilarity measures (DM) or distances have been proposed in the literature which either assess similarity at a specific scale e.g. local or global, or provide a more holistic multi-scale comparison. See references \cite{donnat2018tracking,faust2002comparing,wills2020metrics} for a comprehensive review. While there is a rich body of literature for graph dissimilarity measures (GDM), analogous notions for hypergraph are lacking in literature. To address this gap, we propose two new approaches for hypergraph comparison. Just like for GDMs within each of these HDM approaches we present a collection of DMs which either assess hypergraph similarity at a specific scale or provide a multi-scale comparison. Specifically, key contributions of this paper are as follows:
\begin{itemize}
  \item We develop an indirect approach for comparing hypergraphs by first transforming the hypergraph into a graph and then invoking standard GDMs. In particular we explore clique and star expansion for this transformation. While information about hypergraph structure may be lost during such transformations/projections, the assumption is that relevant salient features may still be preserved which are sufficient to capture key differences between underlying hypergraphs. We refer to these DMs as indirect HDMs.
  \item We introduce another direct approach which relies on tensor based representation of hypergraph which intrinsically captures the multi-way relations encoded by hyperedges. In particular we use adjacency tensor and Laplacian tensor associated with hypergraphs, and tensor algebraic notions of tensor eigenvalues/eigenvectors and higher order singular values to develop new notions of DMs for hypergraphs. We refer to these DMs as direct HDMs.
  \item We test the proposed HDMs on synthetic hypergraphs to assess their usability in discerning between common hypergraph topologies. We also apply the methods to real-world hypergraphs arising in biological datasets.
\end{itemize}

The paper is organized into seven sections. We introduce basic notation and mathematical preliminaries related to hypergraphs, and discuss some desirable characteristics of HDMs in Section \ref{sec:prelim}. In Section \ref{sec:distances}, we provide a short survey of different GDMs. We then use these GDMs in Section \ref{sec:iHDM} to define indirect HDMs based on conversion of hypergraphs into graphs. In Section \ref{sec:dHDM}, we develop notions of direct HDMs using tensor based representation of hypergraphs. Applications to synthetic and real-world hypergraph datasets are presented in Section \ref{sec:results}. We discuss pros/cons of indirect and direct HDMs and directions for future research in Section \ref{sec:discussion}, and conclude in Section \ref{sec:conc}.

\section{Preliminaries}\label{sec:prelim}

\textbf{Hypergraph:} Let $\Vs$ be a finite set. A hypergraph $\g$ is a pair $(\Vs,\Es)$ where $\Es\subseteq \mathcal{P}(\Vs)\setminus \{\emptyset\}$, the power set of $\Vs$. The elements of $\Vs$ are called the vertices, and the elements of $\Es$ are called the hyperedges. We note that in this definition of hypergraph we do not allow for repeated vertices within an hyperedge (often called hyperloops). For a weighted hypergraph, there is positive weight function $w:\Es\rightarrow (0,\infty)$ which defines a weight $w(e)>0$ associated with each hyperedge $e\in \Es$. The degree $d(v)$ of a vertex $v\in\Vs$ is $d(v)=\sum_{e\in \Es|v\in e} w(e)$. The degree of an hyperedge $e$ is denoted by $d(e) = |e|$, where $|\cdot|$ denotes set cardinality. For $k$-uniform hypergraphs, the degree of each hyperedge is the same, i.e. $d(e) = k$.  The vertex-hyperedge incidence matrix $\mH$ is a $|\Vs | \times |\Es|$ matrix where the entry $h(v, e)$ is $1$ if $v \in e$ and $0$ otherwise. By these definitions, we have,
\begin{equation*}
d(v)=\sum_{e\in \Es} w(e)h(v,e),\qquad d(e)=\sum_{v\in \Vs} h(v,e).
\end{equation*}
Let $\mD_e$ and $\mD_v$ be the diagonal matrices consisting of hyperedge and vertex degrees as diagonal entries, respectively. Similarly we will denote by $\mW$ the diagonal matrix formed by hyperedge weights $w(\cdot)$ as its diagonal entries.

Note that a standard graph is a 2-uniform hypergraph. We will denote a standard \textbf{graph} by $\gr$, and by $\mA$ as its adjacency matrix which is $|\Vs|\times |\Vs|$ matrix with entry $(u,v)$ equal to the edge weight $w(e)$ (where $e$ is such that $(u,v)\in e$) if they are connected, and $0$ otherwise. The incidence matrix $\mH$, and the diagonal matrices $\mD_v$ and $\mW$ are similar to as defined above.

\textbf{Hypergraph Dissimilarity Measure (HDM):} Let $\Gs$ be the space of hypegraphs with finite number of vertices. A hypergraph dissimilarity measure (HDM) $\dmh$ is a symmetric non-negative function $\dmh:\Gs\times \Gs \rightarrow [0,\infty)$, i.e. $\dmh(\g,\tg)=\dmh(\tg,\g)$ for any $\g,\tg\in \Gs$.  $\dmh$ quantifies distance between two hypergraphs with larger values indicating higher degree of dissimilarity.

Note that in general $\dmh$ is not required to satisfy $\dmh(\g,\tg)=0$ even when $\g$ and $\tg$ are isomorphic, or the triangular inequality, and thus may not be a valid metric. But depending on the application, such requirements may be further imposed on $\dmh$. Approaches to graph comparison can be roughly divided into two groups, those that consider or require two graphs to be defined on the same set of vertices, and those that do not.

To distinguish DMs which have been defined specifically for graphs in the literature, we will refer to them as \textbf{graph dissimilarity measures} (GDMs), and denote them by $\dm$.


\subsection{Characteristic of Dissimilarity Measures}
In this section we summarize some desirable properties of graph dissimilarity measures which have been noted in the literature, and can also be applied in context of hypergraphs. These properties can serve as guidelines for selecting appropriate DM, and can further be modified and enriched by the data analyst depending on the application at hand. Examples of some desirable properties for the DMs include \cite{koutra2013deltacon}:
\begin{itemize}
  \item  Edge-importance: modifications of the graph structure yielding disconnected components should be penalized more.
  \item Edge-submodularity: a specific change is more important in a graph with a few edges than in a denser graph on the same vertices.
  \item Weight awareness: the impact on the similarity measure increases with the weight of the modified edge.
  \item Focus awareness: random changes in graphs are less important than targeted changes of the same extent.
\end{itemize}
Depending on the application, additional invariance  properties may be imposed on the DMs, such as \cite{tsitsulin2018netlsd}:
\begin{itemize}
  \item Permutation-invariance: implies that if two graphs' structure are the same (i.e., if the two graphs are isomorphic) the DM between them is zero.
  \item Scale-adaptivity: implies that the DM accounts for differences in both local (edge and node) and global (community) features. Using local features only, a DM would deem two graphs sharing local patterns to have near-zero distance although their global properties (such a page-rank features) may differ, and, in reverse, relying on global features only would miss the differences in local structure (such as edge distributions).
  \item Size-invariance: is the capacity of DM to discern that two graphs represent the same phenomenon at a different magnitudes (e.g., two criminal circles of similar structures but different sizes should have near-zero DM). Size-invariance postulates that if two graphs originate from the sampling of the same domain, they should be deemed similar.
\end{itemize}

\section{Review of Graph Dissimilarity Measures}\label{sec:distances}
We review some key GDMs, the material is taken from the survey articles \cite{donnat2018tracking,faust2002comparing,wills2020metrics}. Approaches to graph comparison can be categorized from different perspectives.

One categorization is based on whether the graph comparison method requires the two graphs to be defined on the same set of vertices or not. The former eliminates the need to discover a mapping between node sets, making comparison relatively easier.  A common approach for comparison without assuming node correspondence is to build the DM using graph invariants. Graph invariants are properties of a graph that hold for all isomorphs of the graph. Using an invariant mitigates any concerns with the encoding of the graphs, and the DM is instead focused completely on the graph topology.

Another categorization of graph comparison methods is based on scale at which they compare structures \cite{donnat2018tracking}. Local DMs are only sensitive to differences in direct neighbourhood of each node, while global DMs may ignore node identities and perceive differences only in global structures in the graph such as hubs, communities, number of spanning trees, etc. On the other hand, mesoscopic DMs work at intermediate scale such that they not only preserve vertex identities but also incorporate information characterising vertices by their relationship to the whole graph, rather than uniquely with respect to their neighbours . Finally, multi-scale DMs attempt to capture aspects from multiple scales i.e. local, global and/or mesoscopic in quantifying differences between graphs.

Let $\gr$ and $\tgr$ be two graphs under comparison with adjacency matrices $\mA$ and $\tmA$, respectively, and let their graph Laplacians be $\mL$ and $\tmL$, respectively. The graph Laplacian $\mL$  (and similarly $\tmL$) could be the standard combinatorial Laplacian
\begin{equation}\label{eq:gLun}
\mL_{un}=\mD_v-\mA,
\end{equation}
or its normalized version,
\begin{equation}\label{eq:gLn}
\mL=\mI-\mD_v^{-1/2}\mA\mD_v^{-1/2}.
\end{equation}
We shall denote Laplacian eigenvalues as $0=\lambda_1\leq \lambda_1\leq\cdots\leq \lambda_{\Nnodes}$. The literature remains divided on which version of the Laplacian to pick for defining the DM. Since the eigenvalues of the normalized Laplacian are bounded between 0 and 2, it makes it a more stable and preferable representation. Therefore, if otherwise stated, we will always use the normalized Laplacian $\mL$ in definition of the GDMs.

\begin{itemize}
  \item \textbf{Structural DMs:} The simplest GDMs are obtained by directly computing the difference of the adjacency matrices of the two graphs and then using a suitable norm e.g., Euclidean, Manhattan, Canberra, or Jaccard.  Examples of such GDMs include, the Hamming distance,
    \begin{equation*}
    \dm_H(\gr,\tgr)=\frac{||\mA-\tmA||_1}{\Nnodes(\Nnodes-1)},
    \end{equation*}
  and, the Jaccard distance,
    \begin{equation*}
    \dm_J(\gr,\tgr)=1-\frac{\sum_{ij}\min(\mA_{ij},\tmA_{ij})}{\sum_{ij}\max(\mA_{ij},\tmA_{ij})}.
    \end{equation*}
    Structural DMs focus on differences in the direct local neighborhood of each node, and are agnostic to other more global structures in the graph.

 \item \textbf{Feature-based DMs:} Another possible method for comparing graphs is to look at specific “features” of the graph, such as the degree distribution, betweenness centrality distribution, diameter, number of triangles, number of k-cliques, etc. For graph features that are vector-valued (such as degree distribution) one might also consider the vector as an empirical distribution and take as graph features the sample moments (or quantiles, or other statistical properties). A feature-based distance is a distance that uses comparison of such features to compare graphs.  If we are using node dependent features, the method aggregates a feature-vertex matrix of size $k \times \Nnodes$, where $k$ is number of features selected. This feature-vertex matrix for the two graphs can then be directly compared, or can be further reduced to a ``signature vector" that consists of the mean, median, standard deviation, skewness, and kurtosis of each feature across vertices. These signature vectors are then compared in order to obtain a DM between graphs. NETSIMILE \cite{berlingerio2012netsimile} is an example of feature based distance which uses local and egonet-based features (e.g., degree, volume of egonet as fraction of maximum possible volume, etc.). In the neuroscience literature, in particular, feature-based methods are fairly popular.

In this paper, we will utilize node centrality vector $\mathbf{c}=(c_1,\cdots,c_\Nnodes)^T$ as the feature for graph comparison. Let $c_i,i=1,\cdots,\Nnodes$ and $\tilde{c}_i,i=1,\cdots,\Nnodes$ be normalized (i.e. $|\mathbf{c}_1|=|\mathbf{c}_2|=1$) node centralities for the graphs $\gr$ and $\tgr$, respectively, then centrality based DM is given by:
     \begin{equation}\label{eq:dcen}
      \dm_{C}(\gr,\tgr)=\frac{1}{\Nnodes}\sum_{i=1}^{\Nnodes} |c_i-\tilde{c}_i|.
      \end{equation}
Note that one could use any notion of centrality, e.g. betweeness centrality, closeness centrality, eigenvector centrality etc. as relevant for the application \cite{das2018study}. Since, centrality measures typically characterize vertices as either belonging to the core or to the periphery of the graph, and thus encode global topological information on the status of vertices within the graph, DMs based on centrality capture mesoscopic differences between graphs \cite{donnat2018tracking}.

  \item \textbf{Spectral DMs:} Spectral DMs on the other hand are more suitable for analyses where the critical information in the graph structure is contained at a global scale, rather than locally. Spectral DMs are global measures defined using the eigenvalues of either the adjacency matrix $\mA$ or of some version of the Laplacian $\mL$ \cite{donnat2018tracking}. Both the eigenvalues of the Laplacian and those of the adjacency matrix can be related to physical properties of a graph, and can thus be considered as characteristics of its states. The adjacency matrix does not downweight any changes and treats all vertices equivalently. On the other hand, the eigenspectrum of the Laplacian accounts for the degree of the vertices and is known to be robust to most perturbations. Specific example of spectral DMs include $l_p$ distance on space of Laplacian eigenvalues,
      \begin{equation}\label{eq:deif}
      \dm_{\lambda}(\gr,\tgr)=\frac{1}{\Nnodes}\sum_{i=1}^{\Nnodes} |\lam_i-\tlam_i|^p,
      \end{equation}
      and, spanning tree DM,
      \begin{equation}\label{eq:dspan}
      \dm_{ST}(\gr,\tgr)=|\log(T_{\gr})-\log(T_{\tgr})|,
      \end{equation}
      where, $T_{\gr}$ is number of spanning trees in the graph, given by
      \begin{equation*}
      T_{\gr}=\frac{1}{\Nnodes}\prod_{i=1}^{\Nnodes-1}\lam_i.
      \end{equation*}
      Other DMs include distances based on the eigenspectrum distributions,
      \begin{equation}\label{eq:ddis}
      \dm_{\rho}(\gr,\tgr)=\int |\rho_{\gr}(x)-\rho_{\tgr}(x)|dx,
      \end{equation}
      where,
      \begin{equation*}
      \rho_{\gr}(x)=\frac{1}{\Nnodes}\sum_{i=0}^{\Nnodes-1}\frac{1}{\sqrt{2\pi\sigma^2}}e^{-\frac{(x-\lambda_i)^2}{2\sigma^2}}
      \end{equation*}
      Another related DM is the Ipsen–Mikhailov distance which characterizes the difference between two graphs by comparing their spectral densities, rather than the raw eigenvalues themselves.

     \item \textbf{DELTACON:} This DM is based on the fast belief propagation method of measuring node affinities \cite{koutra2013deltacon}. It uses the fast belief propagation matrix
      \begin{equation*}
      \mathbf{S}=[\mI+\epsilon^2 \mD -\epsilon \mA]^{-1},
      \end{equation*}
      and compares the two representations $\mathbf{S}$ and $\tilde{\mathbf{S}}$ via the Matusita difference, leading to
      \begin{equation} \label{eq:deltacon}
      \mD_{\Delta}(\gr,\tgr)=\left( \sum_{ij} \left(\sqrt{S_{ij}}-\sqrt{\tilde{S}_{ij}}\right)^2\right)^{1/2}.
      \end{equation}
      Fast belief propagation is designed to model the diffusion of information throughout a graph, and so in theory should be able to perceive differences in both global and local structures in the graph.

      \item \textbf{Heat Spectral Wavelets:} An alternative is to derive characterizations of each node’s topological properties through a signal processing approach. A specific example for this type of DM include the heat spectral wavelets in which the eigenvalues are modulated and combined with their respective eigenvectors to yield a “filtered” representation of the graph’s signal. For a given scale factor $\tau>0$ is a scale, structural signature $\mathbf{\xi}_u$ for each node $u$ is defined to be a vector of coefficients,
          \begin{equation*}
            \mathbf{\xi}_u^\tau=(\Psi^\tau_{1,u},\Psi^\tau_{2,u},\cdots,\Psi^\tau_{\Nnodes,u})^T,
          \end{equation*}
          where,
         \begin{equation*}
            \Psi_{v,u}^\tau=\sum_{i=0}^{\Nnodes-1} e^{-\tau \lambda_i}V_{ui}V_{vi},
          \end{equation*}
          with, $\mL=\mV \mLam \mV^T$ being the Laplacian's eigenvector decomposition. Let $\mathbf{\xi}_u=((\mathbf{\xi}_u^{\tau_1})^T,\cdots,(\mathbf{\xi}_u^{\tau_m})^T)^T$ be the combined vector for a set of selected scales $\tau_1,\tau_2,\cdots,\tau_m$. By choosing these scales appropriately, one can capture information on the connectedness and centrality of each node within the network, thereby providing a way to encompass in a single Euclidean vector all the necessary information to characterize vertices' topological status within the graph. Then the heat kernel DM between the graphs amounts to the average $l_2$ distance between corresponding node’s structural embedding $\mathbf{\xi}_i$,  i.e.,
          \begin{equation}\label{eq:dHK}
          \dm_{HK}(\gr,\tgr)=\frac{1}{\Nnodes}\sum_{i=1}^{\Nnodes}||\mathbf{\xi}_i -\tilde{\mathbf{\xi}}_i||_2.
          \end{equation}

      \item \textbf{NetLSD:} Similar to heat spectral wavelet, in network Laplacian spectral descriptor (NetLSD) \cite{tsitsulin2018netlsd} a heat kernel is defined as,
      \begin{equation*}
      \mH_\tau=e^{-\tau\mL}=\mV e^{-\mLam \tau} \mV^T,
      \end{equation*}
      along with its heat trace,
      \begin{equation*}
      h_\tau=\mbox{Tr}(\mH_\tau)=\sum_{j=1}^{\Nnodes}e^{-\lambda_j \tau}.
      \end{equation*}
      Then the NetLSD condenses the graph representation in form of a heat trace signature  $h(\gr)=\{h_\tau\}_{\tau>0}$ which comprises of a collection of heat traces at different time scales. The continuous-time function $h_\tau$ is finally transformed into a finite-dimensional vector by sampling over a suitable time interval. The DM between $\gr$ and $\tgr$ is then taken to be the $l_{\infty}$ norm of vector difference between $h(\gr)$ and $h(\tgr)$.

      Note, that the heat kernel can be seen as continuous-time random walk propagation (where, $(\mH_\tau)_{ij}$ is the heat transferred from node $i$ to node $j$ at time $\tau$), and its diagonal (sometimes referred to as the autodiffusivity function or the heat kernel signature) can be seen as a continuous-time PageRank. As $\tau$ approaches zero, the Taylor expansion yields $\mH_\tau=1-\mL \tau$ meaning the heat kernel depicts local connectivity. On the other hand for large $\tau$,
      $\mH_\tau=1-e^{-\tau\lambda_2}\evec_2\evec_2^T$ where $\evec_2$ is the Fielder vector used in spectral graph clustering, it encodes global connectivity. Thus, the heat kernel localizes around its diagonal, and the degree of localization (as captured by the heat trace) depends on the scale $\tau$, it can thereby be tuned to capture both local and global graph structures.


    \item \textbf{Graph Embedding based DMs:} Given the diversity of structural features in graphs, and the difficulty of designing by hand the set of features that optimizes the graph embedding, several researchers have proposed recently to learn the embedding from massive datasets of existing networks. Such algorithms learn an embedding from a set of graphs into Euclidean space, and then compute a notion of similarity between the embedded graphs \cite{goyal2018graph}. All these approaches rely on the extension of convolutional neural networks to non Euclidean structures, such as manifolds and graphs.

    \item \textbf{Graph Kernels based DMs:} A popular approach to learning with graph-structured data is to make use of graph kernels—functions which measure the similarity between graphs \cite{kriege2020survey}. These kernels can be used for comparing graphs. Many different graph kernels have been defined, which focus on different types of substructures in graphs, such as random walks, shortest paths, subtrees, and cycles. One particular approach is based on graphlets which are small, connected, non-isomorphic, induced subgraph of a larger graph. There are 30 graphlets with 2- to 5- vertices. Each graphlet contains ``symmetrical vertices” which are said to belong to the same automorphism orbit. The automorphism orbits represent topologically different ways in which a graphlet can touch a node. The Graphlet Degree Vector (GDV) of a node generalises the notion of a node’s degree into a 73-dimensional vector where each of the $73$ components of that vector captures the number of times node $n$ is touched by a graphlet at orbit $i$. Using GDV one can then define several different GDMs including: relative graphlet frequency distance, graphlet degree distribution agreement, and graphlet correlation matrix / distance.

\end{itemize}

\section{Approach I: Indirect HDMs Based on Graph Representation}\label{sec:iHDM}
The first approach we a propose for defining HDMs is based on transforming the hypergraph into a graph representation and then invoking the standard GDMs.

\subsection{Graph-based Hypergraph Representation}
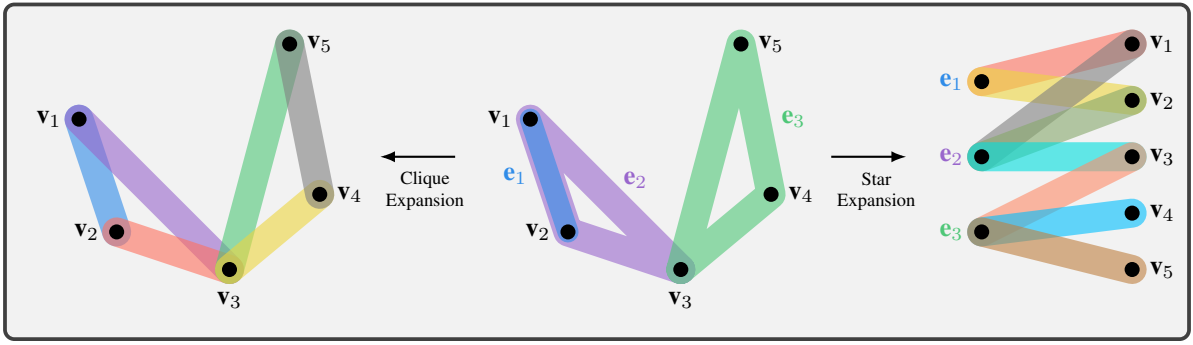
\begin{figure*}[htbp]
\centering
\tcbox[colback=mygray,top=5pt,left=5pt,right=-5pt,bottom=5pt]{
\begin{tikzpicture}
\draw[line width=11pt,line cap=round, color=amethyst,opacity=.6,fill opacity=.5,line join=round] (0,0) -- (-1.5,0.5) -- (-2,2) -- (0,0);
\draw[line width=8pt,line cap=round, color=bleudefrance,opacity=.6,fill opacity=.5,line join=round] (-1.5,0.5) -- (-2,2);
\draw[line width=11pt,line cap=round, color=emerald,opacity=.6,fill opacity=.5,line join=round] (0,0) -- (1.2,1) -- (0.8,3) -- (0,0);

\redpt{0,0}
\redpt{-1.5,0.5}
\redpt{-2,2}
\redpt{1.2,1}
\redpt{0.8,3}

\draw[line width=11pt,line cap=round, color=bleudefrance,opacity=.6,fill opacity=.5,line join=round] (-1.5-6,0.5) -- (-2-6,2);
\draw[line width=11pt,line cap=round, color=amethyst,opacity=.6,fill opacity=.5,line join=round] (-2-6,2) -- (0-6,0);
\draw[line width=11pt,line cap=round, color=bittersweet,opacity=.6,fill opacity=.5,line join=round] (0-6,0) -- (-1.5-6,0.5);
\draw[line width=11pt,line cap=round, color=emerald,opacity=.6,fill opacity=.5,line join=round] (0.8-6,3) -- (0-6,0);
\draw[line width=11pt,line cap=round, color=sandstorm,opacity=.6,fill opacity=.5,line join=round] (0-6,0) -- (1.2-6,1);
\draw[line width=11pt,line cap=round, color=gray,opacity=.6,fill opacity=.5,line join=round] (1.2-6,1) -- (0.8-6,3);

\draw[line width=11pt,line cap=round, color=bittersweet,opacity=.6,fill opacity=.5,line join=round] (4,2.5) -- (6,3);
\draw[line width=11pt,line cap=round, color=sandstorm,opacity=.6,fill opacity=.5,line join=round] (4,2.5) -- (6,2.25);

\draw[line width=11pt,line cap=round, color=gray,opacity=.6,fill opacity=.5,line join=round] (4,1.5) -- (6,3);
\draw[line width=11pt,line cap=round, color=asparagus,opacity=.6,fill opacity=.5,line join=round] (4,1.5) -- (6,2.25);
\draw[line width=11pt,line cap=round, color=blue-green,opacity=.6,fill opacity=.5,line join=round] (4,1.5) -- (6,1.5);

\draw[line width=11pt,line cap=round, color=salmon,opacity=.6,fill opacity=.5,line join=round] (4,0.5) -- (6,1.5);
\draw[line width=11pt,line cap=round, color=capri,opacity=.6,fill opacity=.5,line join=round] (4,0.5) -- (6,0.75);
\draw[line width=11pt,line cap=round, color=brown,opacity=.6,fill opacity=.5,line join=round] (4,0.5) -- (6,0);

\redpt{0-6,0}
\redpt{-1.5-6,0.5}
\redpt{-2-6,2}
\redpt{1.2-6,1}
\redpt{0.8-6,3}

\redpt{0+4,0+0.5}
\redpt{0+4,1+0.5}
\redpt{0+4,2+0.5}

\redpt{0+6,0}
\redpt{0+6,0.75}
\redpt{0+6,1.5}
\redpt{0+6,2.25}
\redpt{0+6,3}

\node at (-2.4, 2) {$\textbf{v}_1$};
\node at (-1.9, 0.5) {$\textbf{v}_2$};
\node at (0, -0.4) {$\textbf{v}_3$};
\node at (1.6, 1) {$\textbf{v}_4$};
\node at (1.2, 3) {$\textbf{v}_5$};

\node at (-2.4-6, 2) {$\textbf{v}_1$};
\node at (-1.9-6, 0.5) {$\textbf{v}_2$};
\node at (0-6, -0.4) {$\textbf{v}_3$};
\node at (1.6-6, 1) {$\textbf{v}_4$};
\node at (1.2-6, 3) {$\textbf{v}_5$};

\node at (0+3.6,2+0.5) {\textcolor{bleudefrance}{$\textbf{e}_1$}};
\node at (3.6, 1.5) {\textcolor{amethyst}{$\textbf{e}_2$}};
\node at (3.6, 0.5) {\textcolor{emerald}{$\textbf{e}_3$}};

\node at (6.4, 3) {$\textbf{v}_1$};
\node at (6.4, 2.25) {$\textbf{v}_2$};
\node at (6.4, 1.5) {$\textbf{v}_3$};
\node at (6.4, 0.75) {$\textbf{v}_4$};
\node at (6.4, 0) {$\textbf{v}_5$};

\node at (-2.2, 1.25) {\textcolor{bleudefrance}{$\textbf{e}_1$}};
\node at (-0.6, 1.2) {\textcolor{amethyst}{$\textbf{e}_2$}};
\node at (1.5, 2) {\textcolor{emerald}{$\textbf{e}_3$}};

\draw [->, thick,-{Latex[length=2mm]}] (-3,1.5) -- (-4,1.5);
\draw [->, thick,-{Latex[length=2mm]}] (2,1.5) -- (3,1.5);
\node at (-3.4, 1.2) {\scriptsize Clique};
\node at (-3.4, 0.9) {\scriptsize Expansion};

\node at (2.6, 1.2) {\scriptsize Star};
\node at (2.6, 0.9) {\scriptsize Expansion};
\end{tikzpicture}
\hspace{0.2cm}}
\caption{Illustration of hypergraph transformation into a graph by clique and star expansion.}\label{fig:expn}
\end{figure*}

There are two main ways to transform a hypergraph in form of a standard graph: clique expansion and star expansion \cite{agarwal2006higher}, see Figure \ref{fig:expn}. Once hypergraph is represented in form of a standard graph, one can define appropriate adjacency matrix and graph Laplacian. Rather than first transforming hypergraph into a graph, some authors define hypergraph Laplacian directly using analogies from the graph Laplacian. However, it was shown in \cite{agarwal2006higher}, that several of these direct definitions are special cases of clique or star expansion which follows from different ways of deriving edge weights for the transformed graph from hyperedge weights of the hypergraph. Thus, we focus on clique expansion and star expansion.


\paragraph{Clique expansion} The clique expansion algorithm constructs a graph $\gr^c=(\Vs,\Es^c \subset \Vs^2)$ from the original hypergraph $\g=(\Vs,\Es)$  by replacing each hyperedge $e = (u_1,\cdots, u_{d(e)}) \in \Es$ with an edge for each pair of vertices in the hyperedge: $\Es^c = \{(u, v) : u, v \in e, e \in \Es\}$. Note that the vertices in hyperedge $e$ form a clique in the graph $\gr^c$. The edge weight $w^c(u,v)$  can be defined in different ways leading to different clique expansions. Thus, the normalized Laplacian of the constructed graph $\g^c$ becomes
\begin{equation*}
\mL^c=\mI-(\mD_v^c)^{-1/2}\mA^c(\mD_v^c)^{-1/2},
\end{equation*}
where, $\mA^c$ is the adjacency matrix
\begin{equation*}
[\mA^c]_{uv}=w^c(u,v),
\end{equation*}
and, $\mD_v^c$ is the vertex degree matrix with diagonal entries $d^c(u), u\in \Vs$.

The \textbf{standard clique} approach minimizes the difference between the edge weight of $\gr^c$ and the weight of each hyperedge $e$ that contains both $u$ and $v$ leading to,
\begin{eqnarray*}
  w^{cs}(u,v)&=& \sum_{e} h(u,e)h(v,e)w(e), \\
  d^{cs}(u) &=& \sum_{e\in\Es} h(u,e)(d(e)-1)w(e),
\end{eqnarray*}
and thus,
\begin{eqnarray}
\mA^{cs}&=&\mH\mW\mH^T,\label{eq:cliquestandA}\\
\mL^{cs}&=&\mI-(\mD_v^{cs})^{-1/2}\mA^{cs}(\mD_v^{cs})^{-1/2} \label{eq:cliquestandL}.
\end{eqnarray}
Choosing weight matrix \cite{bolla1993spectra},
\begin{equation*}
\mW^{co}=\mH \mD_e^{-1} \mH^T,
\end{equation*}
and using combinatorial Laplacian for $\gr^c$, leads to Bolla's Laplacian,
\begin{equation}\label{eq:LB}
\mL^{co}=\mD_v-\mH\mD_e^{-1}\mH^T.
\end{equation}
For an unweighted hypergraph $\g$ i.e. $w(e)=1,\forall e\in \Es$, other choices of weight matrix have been considered, see \cite{agarwal2006higher} for details.

\paragraph{Star expansion} The star expansion algorithm constructs a graph $\gr^*=(\Vs^*,\Es^*)$ from hypergraph $\g=(\Vs,\Es)$ by introducing a new vertex for every hyperedge $e\in \Es$, thus $\Vs^*=\Vs\cup \Es$. It connects the new graph vertex $e$ to each vertex in the hyperedge to it, i.e. $\Es^*=\{(u, e) : u \in e, e \in E\}$. Note that each hyperedge in $\Es$ corresponds to a star in the graph $\gr^*$, and $\gr^*$ is a bi-partite graph. As in clique expansion, different choices can be made for edge weights $w^*(u,e)$ of $\gr^*$.  In general, the adjacency matrix $\mA^*$  of $\gr^*$ can be expressed as,
\begin{equation*}
\mA^*=\left(
        \begin{array}{cc}
          \mZ_{|\Vs|} & \mW^* \\
          (\mW^*)^T & \mZ_{|\Es|} \\
        \end{array}
      \right),
\end{equation*}
and, the normalized Laplacian can be shown to be,
\begin{equation*}
\mL^*=\left(
        \begin{array}{cc}
          \mI & -\mB^* \\
          -(\mB^*)^T & \mI \\
        \end{array}
      \right),
\end{equation*}
where, $\mB^*$ is the $|\Vs|\times |\Es|$ matrix
\begin{equation*}
\mB^*=(\mD_v^*)^{-1/2}\mW^*(\mD_e^*)^{-1/2},
\end{equation*}
where, $\mD_v^*$ and $\mD_e^*$ are degree matrices with diagonal entries $d^*(u)$, and $d^*(e)$, respectively, where,
\begin{eqnarray*}
  d^*(u) &=& \sum_{e\in \Es} w^*(u,e),\qquad u\in \Vs, \\
  d^*(e) &=& \sum_{u\in \Vs} w^*(u,e),\qquad e \in \Es.
\end{eqnarray*}
Note that since number of hyperedges i.e. $|\Es|$ can be large, the star expansion would result in a graph $\gr^*$ which can have very large number of vertices making the application of GDM challenging. Furthermore, even if two hypergraphs $\g$ and $\tg$ are defined on same node set $\Vs$ to begin with, the star expansions, $\gr^*$ and $\tgr^*$, respectively will in general have a different set of vertices.

To alleviate these issues, we propose to use the notion of projected Laplacian as defined in \cite{agarwal2006higher}.  Note that any for a $|\Vs|+|\Es|$ eigenvector $\evec^T=[\evec_v^T, \evec_e^T]$ of $\mL^*$ that satisfies $\mL^*\evec=\lam\evec$, then,
\begin{equation*}
\mB^*(\mB^*)^T\evec_v=(\lam-1)^2\evec_v.
\end{equation*}
Thus, the $|\Vs|$ elements of the eigenvectors of $\mL^*$ corresponding to vertices $\Vs\subset \Vs^*$ are eigenvector of the $|\Vs|\times |\Vs|$ matrix $\mB^*(\mB^*)^T$,
\begin{equation*}
\mB^*(\mB^*)^T=(\mD_v^*)^{-1/2}\mW^*(\mD_e^*)^{-1}(\mW^*)^T(\mD_v^*)^{-1/2}.
\end{equation*}
Given this relationship between $\mB^*(\mB^*)^T$ and $\mL^*$,
\begin{equation*}
\mL^*_p=\mI-\mB^*(\mB^*)^T=\mI-(\mD_v^*)^{-1/2}\mA^*_p(\mD_v^*)^{-1/2},
\end{equation*}
can be considered a projected normalized Laplacian on node set of original graph $\g$, with $\mA^*_p$,
\begin{equation*}
\mA^*_p=\mW^*(\mD_e^*)^{-1}(\mW^*)^T,
\end{equation*}
being the projected adjacency matrix. Note that the eigenvalues of $\mL^*_p$ lie in $[0,1]$, i.e. $0=\lambda_{0p}\leq \cdots, \lambda_{\Nnodes p}\leq 1$.

We next discuss different choices for weights $w^*(u,e)$. The \textbf{standard star} expansion approach assigns the scaled hyperedge weight i.e. $w^{*s}(u,e)=\frac{w(e)}{d(e)}$ to each corresponding graph edge, so that the weight matrix becomes,
\begin{equation*}
\mW^{*s}=\mH\mW\mD_e^{-1},
\end{equation*}
leading to
\begin{eqnarray*}
  d^*(u) &=& \sum_{e\in \Es} h(u,e)w(e)/\delta(e),\quad u\in \Vs, \\
  d^*(e) &=& w(e),\quad e \in \Es,
\end{eqnarray*}
expressed in terms of original hypergraph's $\g$ properties.
Another choice is $w^*(u,e)=w(e)$, i.e.,
\begin{equation*}
\mW^{*z}=\mH\mW,
\end{equation*}
which leads to $\mD_v^*=\mD_v$ and $\mD_e^*=\mW\mD_e$, and thus resulting in,
\begin{equation}\label{eq:ZA}
\mA^{*z}_p=\mH\mW\mD_e^{-1}\mW\mH^T,
\end{equation}
and
\begin{equation}\label{eq:ZL}
\mL^{*z}_p=\mI-\mD_v^{-1/2}\mH\mW\mD_e^{-1}\mH^T\mD_v^{-1/2},
\end{equation}
which is the same hypergraph Laplacian as the one proposed by Zhuo et. al. \cite{zhou2007learning}.  This definition of hypergraph Laplacian originates from relaxation of normalized hypergraph cut problem analogous to the standard normalized graph cut problem. Infact, the eigenvector of $\mL^{*z}$ corresponding to its second smallest eigenvalue encodes the information about subsets of vertices in the hypergraph which are weakly connected to each other.

In whatever follows, we will use the standard clique expansion (Eqns. \ref{eq:cliquestandA} and \ref{eq:cliquestandL}), and projected star expansion based on Zhuo et. al. construction (Equn.s \ref{eq:ZA} and \ref{eq:ZL}) for transforming hypergraph into graph for dissimilarity comparison using GDMs.

\subsection{Indirect HDMs}
Let $\g$ be a hypergraph, and let $\gr_H$ be the graph obtained by one of approaches discussed in the previous section. We will denote this transformation as $\gr_H=\mcT(\g)$. Given a transformation $\mcT$ and a GDM $\dm$, an indirect HDMs $\dmh$ induced by the pair $(\mcT,\dm)$ is given by,
\begin{equation}\label{eq:inDHM}
\dmh_{\mcT,\dm}(\g,\tg)\equiv \dm(\mcT(\g),\mcT(\tg))=\dm(\gr_H,\tgr_H).
\end{equation}
Note that depending on whether $\g$ and $\tg$ have known or unknown node correspondence, appropriate $\dm$ can be chosen from the GDMs discussed in Section \ref{sec:distances} or any other available in literature. Furthermore, depending on application one can pick $\dm$ to capture local, global, mesoscopic or multi-scale differences.

\section{Approach II: Direct HDMs Based on Tensor Based Hypergraph Representation}\label{sec:dHDM}
In this section we propose a second approach for defining HDMs which is based on hypergraph representations that intrinsically capture multi-way relations using tensors.

\subsection{Tensor Preliminaries}
A tensor is a multidimensional array \cite{doi:10.1137/07070111X, Kolda06multilinearoperators, chen2019multilinear,chen2021multilinear}. The order of a tensor is the number of its dimensions, and each dimension is called a mode. An $m$-th order real valued tensor will be denoted by $\textsf{X}\in \mathbb{R}^{J_1\times J_2\times  \dots \times J_m}$, where $J_k$ is the size of its $k-$th mode.

The inner product of two tensors $\textsf{X},\textsf{Y}\in \mathbb{R}^{J_1\times J_2\times \dots \times J_m}$ is defined as,
\begin{equation*}
\langle \textsf{X},\textsf{Y}\rangle =\sum_{j_1=1}^{J_1}\dots \sum_{j_m=1}^{J_m}\textsf{X}_{j_1j_2\dots j_m}\textsf{Y}_{j_1j_2\dots j_m},
\end{equation*}
leading to the tensor Frobenius norm $\|\textsf{X}\|^2=\langle \textsf{X},\textsf{X}\rangle$. We say two tensors $\textsf{X}$ and $\textsf{Y}$ are orthogonal if the inner product $\langle \textsf{X},\textsf{Y}\rangle =0$.
The matrix tensor multiplication $\textsf{X} \times_{k} \textbf{A}$ along mode $k$ for a matrix $\textbf{A}\in  \mathbb{R}^{I\times J_k}$ is defined by,
$
(\textsf{X} \times_{k} \textbf{A})_{j_1j_2\dots j_{k-1}ij_{k+1}\dots j_m}=\sum_{j_k=1}^{J_k}\textsf{X}_{j_1j_2\dots j_k\dots j_N}\textbf{A}_{ij_k}.
$
This product can be generalized to what is known as the Tucker product,
\begin{equation}\label{eq5}
\begin{split}
\textsf{X}\times_1 \textbf{A}_1 \times_2\textbf{A}_2\times_3\dots \times_{m}\textbf{A}_m \in  \mathbb{R}^{I_1\times I_2\times\dots \times I_m}.
\end{split}
\end{equation}

Higher-Order Singular Value Decomposition (HOSVD) is a multilinear generalization of matrix SVD to tensors \cite{doi:10.1137/S0895479896305696}. HOSVD of a tensor $\textsf{X}\in\mathbb{R}^{J_1\times J_2\times \dots \times J_m}$ is given by:
\begin{equation}
\textsf{X} = \textsf{S}\times_1 \textbf{U}_1\times_2\dots \times_m \textbf{U}_m,
\end{equation}
where, $\textbf{U}_k\in\mathbb{R}^{J_k\times R_k}$ are orthogonal matrices, and $\textsf{S}\in\mathbb{R}^{R_1\times R_2\times \dots \times R_m}$ is called the core tensor.  The quantity $R_k\leq J_k$ is referred to as the $k$-mode multilinear rank of $\textsf{X}$, and equal to rank of $k-$ mode matrix unfolding of $\textsf{X}$. The subtensors $\textsf{S}_{j_k=\alpha}$ of $\textsf{S}$ obtained by fixing the $k$-th mode to $\alpha$, have the properties:
\begin{enumerate}
\item all-orthogonality: two subtensors $\textsf{S}_{j_k=\alpha}$ and $\textsf{S}_{j_k=\beta}$ are orthogonal for all possible values of $k$, $\alpha$ and $\beta$ subject to $\alpha\neq \beta$;
\item ordering: $\|\textsf{S}_{j_k=1}\|\geq \dots \geq \|\textsf{S}_{j_n=J_n}\|\geq 0$ for all possible values of $k$.
\end{enumerate}
The Frobenius norms $\|\textsf{S}_{j_k=j}\|$, denoted by $\gamma_{j}^{(k)}$, are known as the $k$-mode singular values of $\textsf{X}$.  De Lathauwer et al. \cite{doi:10.1137/S0895479896305696} showed that the number of nonvanishing $k$-mode singular values of a tensor is equal to its $k$-mode multilinear rank, i.e. $R_k$. HOSVD can be computed using a sequence of matrix SVDs, and by introducing SVD truncations yields a quasi-optimal solution to the low mutilinear rank approximation problem.

Consider a $m-$th order $n$ dimensional cubical (i.e. with equal size $J_i=n,i=1,\cdots,m$ in all modes) tensor $\tA \in \mathbb{R}^{n\times n\times \dots\times n}$. $\tA$ is called supersymmetric if $\tA_{i_1,\cdots,i_m}=\tA_{\sigma(i_1,\cdots,i_m)}$ for all $\sigma \in \Sigma_m$, the symmetric group of $m$ indices. To a $n-$ vector $\vx=(x_1,\cdots,x_n)^T$, real or complex, define a $n$-vector via Tucker product as:
\begin{equation*}
\tA \vx^{m-1}=\tA \times_2 \vx \times_3 \cdots \times_m \vx.
\end{equation*}

There are many different notions of tensor eigenvalues/eigenvectors \cite{qi2005eigenvalues,lim2005singular}.
A pair $(\lambda,\vx) \in \R \times \{\R^n \setminus \{0\}\}$ is called
\begin{itemize}
  \item H-eigenvalue/eigenvector (or H-eigenpair)  of $\tA$ if they satisfy,
\begin{equation}
\tA \vx^{m-1}=\lambda \vx^{[m-1]},
\end{equation}
where, $(\vx^{[m-1]})_i=x_i^{m-1}$ .
  \item Z-eigenvalue/eigenvector (or Z-eigenpair) of $\tA$ if they satisfy,
\begin{eqnarray}
\tA \vx^{m-1}&=&\lambda \vx,\notag\\
x_1^2+\cdots+x_n^2&=&1.
\end{eqnarray}
  \item $l^p$-eigenvalue/eigenvector (or $l^p$-eigenpair) of $\tA$  for any $p>0$ if they satisfy,
  \begin{eqnarray}
\tA \vx^{m-1}&=&\lambda \vx^{[p-1]},\notag\\
x_1^p+\cdots+x_n^p&=&1.\label{eq:lpeig}
\end{eqnarray}
\end{itemize}
Note that $l^p$-eigenvalue/eigenvector reduce to H-eigenvalue/eigenvector and Z-eigenvalue/eigenvector for $p=m$ and $p=2$, respectively, where note the constraint in (\ref{eq:lpeig}) is superfluous for $p=m$.

It was proved in \cite{qi2005eigenvalues} that H-eigenvalues and Z-eigenvalues exist for an even order real supersymmetric tensor. A numerical procedure for computing $H$ eigenvalues is provided in \cite{chen2016computing} with an associated MATLAB toolbox \cite{tenEig}. The procedure involves homotopy continuation type method which can be computationally intensive, thus making it challenging to scale to large order/size tensors.

\subsection{Tensor Based Hypergraph Representation}
We follow tensor based formulation proposed in \cite{banerjee2017spectra} to define hypergraph adjacency tensor and Laplacian tensor. Let $\g=(\Vs,\Es,w(\cdot))$ be a weighted hypergraph with $\Nnodes$ vertices, and $k$ be the maximum cardinality of the hyperedges, i.e.  $k = \max\{|e| : e \in \Es\}$.

The adjacency tensor $\tA\in\mathbb{R}^{\Nnodes\times \Nnodes\times \dots\times \Nnodes}$ of $\g$, which is a $k$-th order $n$-dimensional supersymmetric tensor, is defined as,
\begin{equation}
\tA_{j_1j_2\dots j_k} = \begin{cases} \frac{w(e) s}{\alpha} \text{ if $e=(i_1,i_2,\dots,i_s)\in \Es$}\\ \\0, \text{ otherwise}\end{cases},
\end{equation}
where, $j_1j_2\dots,j_k$ are chosen in all possible ways from $\{i_1,i_2,\dots,i_s\}$ with atleast once for each element of the set, and
\begin{equation*}
\alpha =\sum_{k_1,\cdots,k_s\geq 1, \sum_{i=1}^s k_i=k} \frac{k!}{\prod_{l=1}^s k_i!}.
\end{equation*}
Using the adjacency tensor, the degree $d(v_i)$, of a vertex $v_i \in \Vs$, can be expressed as,
\begin{equation}\label{eq:degHyper}
d(v_i)=\sum_{j_1j_2j_{k-1}=1}^\Nnodes \tA_{ij_1j_2j_{k-1}}.
\end{equation}
The choice of the nonzero coefficients $\frac{w(e) s}{\alpha}$  preserves the degree of each node, i.e., the degree of node $j$ computed using (\ref{eq:degHyper}) with weights as defined above is equal to number of hyperedges containing the node in the original non-uniform hypergraph. Note that for a $k$-uniform hypergraph, above definition simplifies to,
\begin{equation}
\tA_{j_1j_2\dots j_k} = \begin{cases} \frac{w(e)}{k-1!} \text{ if $e=(i_1,i_2,\dots,i_k)\in \Es$}\\ \\0, \text{ otherwise}\end{cases}.
\end{equation}

Let $\tD$ be a $k$th-order $\Nnodes$-dimension super-diagonal tensor with nonzero elements $d_{ii\cdots i}=d(v_i)$.  The hypergraph Laplacian tensor is defined as,
\begin{equation}\label{eq:tensorL}
\tL=\tD-\tA,
\end{equation}
which is also a $k$th-order $\Nnodes$-dimension super-symmetric tensor. Similarly, normalized hypergraph Laplacian tensor can also be defined. We recall a result from \cite{banerjee2017spectra}, which establishes following properties (which are analogous to case of graph Laplacian) of $\tL$ ,
\begin{itemize}
  \item $\tL$ has an H-eigenvalue $0$ with eigenvector $\evec=(1,1,\cdot,1)^T\in \R^\Nnodes$. Moreoever, $0$ is the unique $H^{++}$-eigenvalue of $\tL$,
  \item $\Delta$ is the largest $H^+$-eigenvalue of $\tL$, where $\Delta$ is maximum node degree of $\g_H$,
  \item $(d(v_i),\mathbf{e}_j)$ is an H-eigenpair, where $\mathbf{e}_j\in \R^n$ are the standard basis vectors.
 \end{itemize}
H-eigenvalues of $\tL$, thus encode global structural properties of a hypergraph, and  we propose to use them in generalizing spectral GDM for hypergraphs.

We next discuss HOSVD of $\tL$ and associated properties. Since $\tL$ is supersymmetric, any mode unfolding of $\tL$ would yield the same unfolding matrix with the same singular values which we denote by $\gamma_{j},j=1,\cdots,\Nnodes$ (note that we have removed dependence of $\gamma_{j}^k$ on mode $k$). It was shown in \cite{9119161} that the singular values of $\tL$ encode structural properties of the hypergraph, such as vertex degrees, path lengths, clustering coefficients and nontrivial symmetricity for uniform hypergraphs, and thus can be used to quantify differences in hypergraph structure. Moreover, a fast and memory efficient tensor train decomposition (TTD)-based computational framework was developed in \cite{9119161} to compute the singular values for uniform hypergraphs. Given these two desirable features, we also propose to use singular values of $\tL$ as an alternative to H-eigenvalues in defining spectral HDM.

The notion of centrality has been generalized for hypergraphs. H/Z eigenvectors of the adjacency tensor $\tA$ are used to define hypergraph eigenvector-centrality  \cite{benson2019three}. In particular, the H/Z-eigenvector centrality $\mathbf{c}\in \R^\Nnodes$ is defined to be an H/Z-eigenvector of the adjacency tensor $\tA$ such that it is positive i.e. $\mathbf{c}>0$ with a positive H/Z-eigenvalue, i.e. $\lambda>0$. By Perron-Frobenious theorem for non-negative tensors \cite{chang2008perron}, such positive H/Z-eigenvectors exist under certain irreducibility conditions on $\tA$. While such a positive Z-eigenpair may not be unique, the H-eigenpair is always unique upto scaling. Along similar lines, authors in \cite{tudisco2021node} define node and hyperedge centralities as vectors $\mathbf{c}\in \R^\Nnodes$ and $\mathbf{e}\in \R^m$, respectively, that satisfy,
\begin{eqnarray}
  \mathbf{c}\lambda &=& g(\mH \mW f(\mathbf{e})), \label{eq:cen1}\\
  \mathbf{e}\mu &=& \psi(\mH^T \phi(\mathbf{c})), \label{eq:cen2}\\
  \mbox{s.t.} && \mathbf{c},\mathbf{e}>0, \lambda,\mu>0. \label{eq:cen3}
\end{eqnarray}
In above system of equations, $\mH$ is the incidence matrix and $\mW$ is the hyperedge weight matrix for $\g$ (and we have assumed that all vertices have unit weight consistent with setting in this paper), and $g,f,\phi,\psi:\R^{+}\rightarrow \R^{+}$ are appropriately chosen non-negative functions on non-negative real domain. Furthermore note that these scalar functions are extended on vectors by defining them as mappings that act in a componentwise fashion. By invoking Perron-Frobenious theorem for multi-homogeneous mappings \cite{gautier2019unifying}, it was proved that under certain conditions on the scalar functions,  the solution of above system exists and is unique. A nonlinear power method with convergence guarantees is proposed to solve the above system. Furthermore, it is shown that with choices $f(\mathbf{x})=\mathbf{x}, g(\mathbf{x})=\mathbf{x}^{1/(p+1)}, \psi(\mathbf{x})=e^{\mathbf{x}}$ and $\psi(\mathbf{x})=\ln (\mathbf{x})$ , the node centrality vector $\mathbf{c}$ is also a $l^p$-tensor eigenvector, and thus further generalizing the notion of H/Z-eigenvector centrality.

\subsection{Direct HDMs}
For tensor based representation we define a set of DMs along similar lines as discussed in Section \ref{sec:distances}. Let $\g$ and $\tg$ be two hypergraphs with same node set and same maximum hyperedge cardinality, and let $(\tA,\tL)$ and $(\ttA,\ttL)$ be corresponding adjacency and Laplacian tensor, respectively.
\begin{itemize}
  \item \textbf{Structural HMDs:} It is straightforward to generalize the Hamming and Jaccard distance for graphs to tensor based representation as follows:
    \begin{equation*}
    \dmh_H(\g,\tg)=\frac{||\tA-\ttA||_1}{\mathcal{N}},
    \end{equation*}
    where,
    \begin{equation*}
    ||\tA ||_1=\sum_{i_1=1}^\Nnodes \sum_{i_2=1}^\Nnodes\cdots \sum_{i_k=1}^\Nnodes |\tA_{i_1,\cdots,i_k}|,
    \end{equation*}
     is tensor $1$-norm and $\mathcal{N}=\Nnodes^k-\Nnodes$ is a normalization constant, and
    \begin{equation*}
    \dmh_J(\g,\tg)=1-\frac{\sum_{j_1j_2\dots j_k}\min(\tA_{j_1j_2\dots j_k},\ttA_{j_1j_2\dots j_k})}{\sum_{j_1j_2\dots j_k}\max(\tA_{j_1j_2\dots j_k},\ttA_{j_1j_2\dots j_k})},
    \end{equation*}
    respectively.

  \item \textbf{Feature Based HMDs:} As in feature based GDM, we can use specific “features” of the hypergraph, such as the node degree distribution, different notions of centrality, diameter, etc for use in comparing hypergraphs. If we are using node dependent features, the method aggregates a feature-vertex matrix of size $k \times \Nnodes$, where $k$ is number of features selected. This feature-vertex matrix for the two hypergraphs can then be directly compared, or can be further reduced to a ``signature vector" as in the graph case, and used to obtain a DM between hypergraphs. As in graph case (see Section \ref{sec:distances}) we propose to use tensor based hypergraph centrality as the feature for comparison. Let $c_i,i=1,\cdots,\Nnodes$ and $\tilde{c}_i,i=1,\cdots,\Nnodes$ be normalized (i.e. $|\mathbf{c}|_1=|\tilde{\mathbf{c}}|_1=1$) node centralities for $\g$ and $\tg$, respectively, then centrality based HDM is given by,
      \begin{equation}\label{eq:dtcen}
      \dmh_{C}(\g,\tg)=\frac{1}{\Nnodes}\sum_{i=1}^{\Nnodes} |c_i-\tilde{c}_i|.
      \end{equation}
      While in above definition one could use any notion of hypergraph centrality, we propose to use the node centrality defined by (\ref{eq:cen1})-(\ref{eq:cen3}) in our application.

  \item \textbf{Spectral HMDs:} Let the ordered set of H-eigenvalues of $\tL$ be $\lam_1,\cdots,\lam_{p}$ i.e. $\lam_1\leq \lam_2\cdots \leq \lam_q$ , and similarly let $\tilde{\lam}_1,\cdots,\tilde{\lam}_{\tilde{q}}$ be the ordered set for $\ttL$. Note that in general $q\neq \tilde{q}$. Without loss of generality, assume $\tilde{q}>q$ and define an extended set of H-eigenvalues $\overline{\lam}_1,\cdots,\overline{\lam}_{\tilde{q}}$ for $\g$, where $\overline{\lam}_i=0,i\leq \tilde{q}-q$, and $\overline{\lam}_i=\lam_{i-(\tilde{q}-q)},i=\tilde{q}-q+1,\cdots,\tilde{q}$.
      The $l_p$ distance on space of H-eigenvalues can then be defined as,
      \begin{equation}\label{eq:deifH}
      \dmh_{\lambda}(\g,\tg)=\frac{1}{\tilde{q}}\sum_{i=1}^{\tilde{q}-1} |\lam_i-\tlam_i|^p.
      \end{equation}
      As discussed above, we similarly propose to use higher order singular values, leading to,
      \begin{equation}\label{eq:deifS}
       \dmh_{\gamma}(\g,\tg)=\frac{1}{\Nnodes}\sum_{i=1}^{\Nnodes-1} |\gamma_i-\tilde{\gamma}_i|^p,
      \end{equation}
      where, $\gamma_i,i=1,\cdots,\Nnodes$ and $\tilde{\gamma}_i,i=1,\cdots,\Nnodes$ are higher order singular values of $\tL$ and $\ttL$, respectively. We will refer to $\dmh_{\lambda}$ and $\dmh_{\gamma}$ as spectral-H and spectral-S HDMs, respectively.

      \item \textbf{Hypergraph Embedding based HDMs:} Recently, graph convolutional neural networks have been extended to hypergraphs \cite{payne2019deep,feng2019hypergraph}.  Thus, as for graphs, one can learn an embedding from a set of hypergraphs into Euclidean space, and then compute a distance between the embedded hypergraphs.

    \item \textbf{Hypergraph Kernel based HDMs:} The notion of graph kernels has been generalized to hypergraphs,  see for example \cite{lugo2017classification,bai2021hypergraph}. These kernels can be used for comparing hypergraphs as in the graph case.

\end{itemize}

\section{Numerical Studies}\label{sec:results}
In this section we assess the performance of indirect and direct HDMs on synthetic hypergraphs and real world biological datasets. For these studies we have chosen one representative example of a local, global, mesoscopic and multiscale HDMs, namely, Hamming HDM (local), spectral HDM (global), centrality based HDM (mesoscopic) and deltaCon HDM (multiscale).

\subsection{Synthetic Hypergraphs}
To generate synthetic hypergraphs, we consider three families of generative models: Erdös-Rényi (ER), Barabási-Albert (BA) and Watts-Strogats (WS). These three models are widely used as test-beds in a variety of network science problems and have varying structural complexity. ER model \cite{bollobas2001random} leads to ``structureless" graph in the sense that the statistical properties of each edge and vertex in the graph is exactly same. In BA model \cite{barabasi1999emergence}, on the other hand, the node degree distribution behaves as a power-law due to preferential attachment, and that impacts both its global and local structure. On the local scale, vertices in graph tend to connect exclusively to highest-degree vertices in the graph, rather than to one another, generating a tree-like topology. The high-degree vertices acts like hub which are by definition are global structures as they touch a significant portion of rest of the graph, thereby increasing the connectivity throughout the graph. WS model \cite{watts1998collective} on a global scale looks like an uncorrelated random graph in which it exhibits no communities or high-degree vertices but has small average shortest path length between vertices, while at local scale it shows high clustering compared to the BA model.

The three models, originally developed for graphs, have been generalized to the hypergraph case. We will restrict to procedure of construction of $k-$ uniform hypergraphs in each family. The user specifies the desired number of vertices $\Nnodes$, desired number of hyperedges $m$ and some additional parameters depending on the model as discussed below.

\paragraph{Erdos–Renyi (ERH)} There are $\binom {\Nnodes}{k}$ possible hyperedges in a $k$-uniform hypergraph. To construct a random $k-$ uniform hypergraph, we uniformly sample $m$ hyperedges from this set without repetition.

\paragraph{Scale Free Hypergraph (SFH$(\mu)$)}  To construct an $k$-uniform SFH, we follow the generative model from \cite{jhun2019simplicial}:
\begin{itemize}
  \item [i.] Assign each node a probablity $p_i$ as:
  \begin{equation*}
    p_i=\frac{i^{-\mu}}{\sum_{j=1}^\Nnodes j^{-\mu}},\quad i=1,\cdots,\Nnodes,
  \end{equation*}
  where, $0<\mu<1$ is a user chosen parameter.
  \item [ii.] Select $k-$ distinct vertices with probabilities $p_{i_1},\cdots,p_{i_k}$. If the hypergraph does not already contain a hyperedge of those chosen $k$ vertices, then add the hyperedge to the hypergraph
  \item [iii.] Repeat step ii) $m$ times.
\end{itemize}
Similar to BA graph, this procedure produces a hypergraph with vertices with average degrees $<d>$ having a power-law distribution, $\Prob_k(<d>)\sim <d>^{-\lambda}$ with $\lambda=1+\frac{1}{\mu}$.

\paragraph{Watts-Strogats Hypergraph (WSH$(p)$)} Using a procedure similar for WS graphs, WSH is constructed as follows, \cite{9119161}:
\begin{itemize}
  \item [i.] Construct a d-regular $k$-uniform hypergraph with $n$ vertices, and add extra hyperedges in every $k+1$ vertices. We refer to hyperedges in this hypergraph as the initial hyperedges.
  \item [ii.] Select an initial hyperedge and generate a new hyperedge with $k$ vertices chosen uniformly at random. If the new hyperedge does not exist, with probability $p$, replace the selected hyperedge with the new hyperedge. Here $0<p<1$ is a rewiring probability as specified by the user.
  \item [iii.] Repeat step ii), till all initial hyperedges have been iterated on.
\end{itemize}

\begin{figure}[hbt!]
    \centering
    \tcbox[colback=white,top=5pt,left=1pt,right=-1pt,bottom=5pt]{
    \includegraphics[trim=4cm 6.5cm 3.5cm 7cm,clip,scale=0.55]{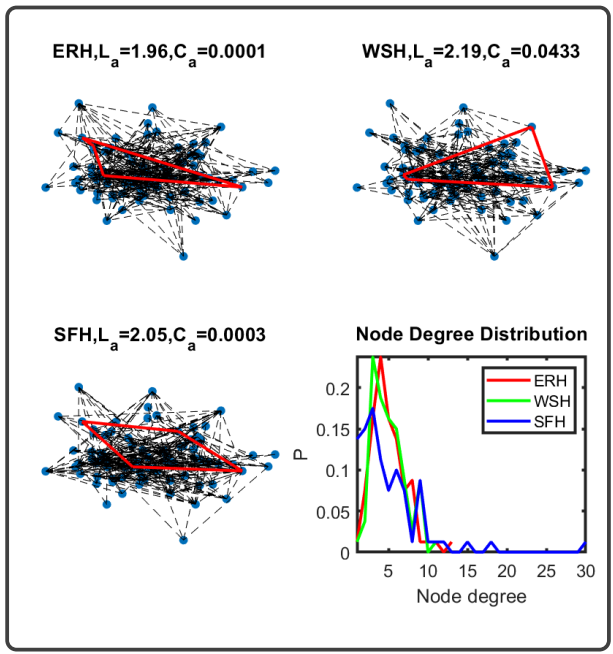}}
    \caption{Examples of $4$-uniform ERH, WSH and SFH with $\Nnodes=80$ and $m=100$. Also shown are node degree distribution, average path length and clustering coefficients for each case. The red polygons represent a selected hyperedge.} \label{fig:syngraphcompare}
\end{figure}

Figure \ref{fig:syngraphcompare}, show a realization of each of these three hypergraph models with $k=4$, $\Nnodes=100$ and $m=125$. To asses if these hypergraph models posses similar structural properties as their graph counterparts, we compare their node degree distribution, average path length, and clustering coefficient. For computing average path length, we use:
\begin{equation}
L_{\text{a}} = \frac{1}{n(n-1)}\sum_{j\neq i}d(v_j,v_i),
\end{equation}
where, $d(v_j,v_i)$ denotes the shortest distance between vertices $v_j$ and $v_i$. Note in computing shortest distance, two vertices are considered adjacent if they share a common hyperedge. For clustering coefficient we use definition from \cite{9119161},
\begin{equation}
\begin{split}
C_j &= \frac{|\{e_{i_1i_2\cdots i_k}: v_{i_1},v_{i_2},\cdots,v_{i_k}\in \Vs_j,e_{i_1i_2\cdots i_k} \in\Es\}|}{\binom {|\Vs_j|}{k}},\\& \Rightarrow C_{\text{a}} =\frac{1}{n}\sum_{j=1}^nC_j,
\end{split}
\end{equation}
where, $\Vs_j$ is the set of vertices that are immediately connected to $v_j$ i.e. share a hyperedge with node $v_j$, and $\binom {|\Vs_j|}{k} = \frac{|\Vs_j|!}{(|\Vs_j|-k)!k!}$ returns the binomial coefficients. If $|\Vs_j|<k$, we set $C_j=0$.

As can be seen from Fig. \ref{fig:syngraphcompare}, the ERH and WSH construction results in a hypergraph with almost homogenous degree distribution. The SFH, on the other hand, shows power-law degree distribution for node degrees, as discussed above. The WSH model shows high clustering coefficient compared to ERH and SFH as expected.

\subsection{Performance on Synthetic Hypergraphs}
We next assess and compare the effectiveness of different HDMs in differentiating hypergraphs with distinct structural features, i.e., originating from the different models. In other words, to yield a good performance, a HDM should be able to assign small distance to hypergraph pairs coming from the same model but large HDM values to pairs coming from different models.

A systematic approach for quantification of the performance can be accomplished via the receiver operating characteristic (ROC) curve \cite{fawcett2006introduction}. The ROC curve is created by plotting the true positive rate (TPR) against the false positive rate (FPR) at various threshold settings. For a given HDM, one defines a threshold $\epsilon> 0$ and classifies two hypergraphs as belonging to the same model class if their HDM value is less than $\epsilon$. Given that the correct classes are known, TPR and FPR values can be computed to quantify the accuracy of classifying all the hypergraphs. The procedure is then repeated by varying $\epsilon$, obtaining the ROC curve which, ideally, should have TPR equal to 1 for any FPR value. Furthermore, one can compute the area under curve (AUC) which is equal to the probability that a classifier will rank a randomly chosen positive instance higher than a randomly chosen negative one. Thus, for a perfect classifier, AUC $= 1$, while for a classifier that randomly assigns observations to classes, AUC$=0.5$.

\begin{figure}[hbt!]
    \centering
    \includegraphics[trim=2cm 7cm 2cm 7cm,clip,scale=0.5]{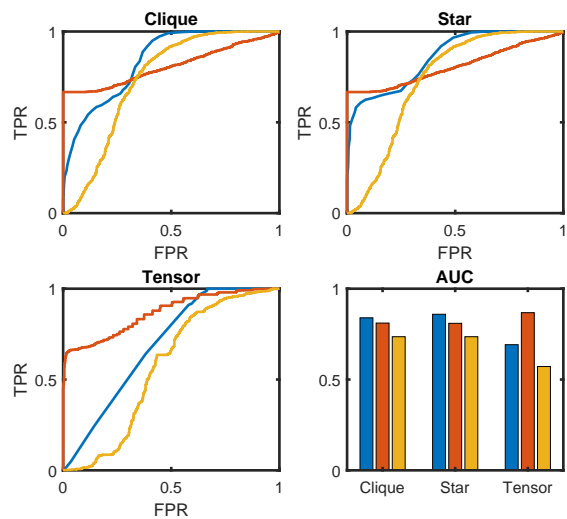}
    \caption{ROC curves and AUC for different HMDs: Hamming (blue), Spectral (red) and Centrality (orange).}
    \label{fig:roc}
\end{figure}

\begin{figure}[hbt!]
    \centering
    \includegraphics[trim=2cm 7.5cm 2cm 7.5cm,clip,scale=0.5]{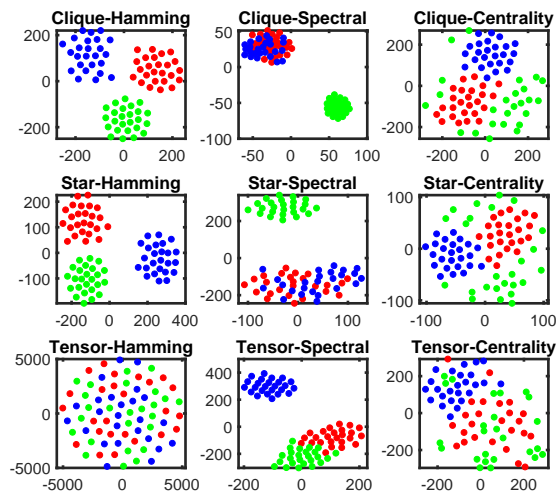}
    \caption{Embedding for different hypergraphs in 2d using tSNE applied to HDMs: ERH (red), WSH (green) and SFH (blue).}
    \label{fig:tsne}
\end{figure}

For testing purposes, we consider hypergraph size $\Nnodes=40$, $m=50$ and generate $25$ networks for each of the three types, resulting in a total population of $N_g=75$ hypergraphs. We then compute all the pairwise HDM using each method, ending up with $N_g \times N_g$ distance matrix for each case. We then generate ROC curves for each HDM considered using procedure discussed above. In addition, to create a visual aid, we also generate a 2d embedding for each hypergraph in the population by applying t-distributed stochastic neighbor embedding (tSNE) \cite{hinton2002stochastic} to the distance matrix corresponding to each HDM. Figure \ref{fig:roc} shows the ROC curves, while the 2d embedding is shown in Figure \ref{fig:tsne}, where we have labeled each point corresponding to different instances of hypergraph using distinct colors based on its known model condition. Note that for an unweighted uniform hypergraph, the adjacency matrices for clique (Eqn. (\ref{eq:cliquestandA})) and star expansion (Eqn. (\ref{eq:ZA})) become same upto a scale factor. Hence, we find that the ROC curves for Hamming and centrality based indirect HDMs are similar.

\subsection{Test on Real Datasets}
In this section we assess the performance of HDMs in grouping sets of real-world networks. In order to assess statistical significance while comparing two hypergraphs, we propose to use the permutation test.

\subsubsection{Permutation Test for HDM}
Consider a hypothesis testing problem:
\begin{itemize}
  \item Null ($H_0$): $\g_1$ and $\g_2$ are similar,
  \item Alternative ($H_1$): $\g_1$ and $\g_2$ are dissimilar.
\end{itemize}
Since the null distribution for $H_0$ is unknown, we use a permutation test \cite{good2006permutation} to empirically estimate it. Let $\dm$ be any of the HDMs, and let $p_s$ be the desired significance level of the test. The steps in the permutation test involve:
\begin{itemize}
  \item Step 1: Randomly generate a family of hypergraphs $\{\g^r_i\}_{i=1}^N$ which are similar to $\g_1$, and compute $\dm_i=\dm(\g_1,\g_i^r),i=1,\cdots,N$.
  \item Step 2: Compute $\dm_{12}=\dm(\g_1,\g_2)$.
  \item Step 3: Compute $p$-value as $p=\frac{1}{N}\sum_{i=1}^N \mathcal{I}_{\dm_{12}}(\dm_i)$, where $\mathcal{I}_x$ is indicator function, i.e. $\mathcal{I}_z(x)=1$ if $x>z$ and $\mathcal{I}_z(x)=0$ otherwise.
  \item Step 4: Reject $H_0$ if $p\leq p_s$.
\end{itemize}
In Step 1 one could use Erdos-Reyni (ER) or Chung–Lu (CL) procedure \cite{aksoy2020hypernetwork} to randomly generate hypergraphs with similar characteristics as $\g_1$. Let  $d^1_v = (d(v_1),\cdots,d(v_\Nnodes))$ and $d^1_e = (d(e_1),\cdots,d(e_m))$ be vertex degree and hyperedge size distribution vectors of $\g_1$. Let $c =\sum_{i=1}^\Nnodes d(v_i)=\sum_{i=1}^m d(e_i)$, and vertex-hyperedge membership probability in $\g_1$ be $p$, where
\begin{equation*}
p=\frac{c}{m\Nnodes}.
\end{equation*}

ER procedure selects vertices uniformly at random for each hyperedge with probability $p$. Thus, for each of the $\Nnodes m$ vertex-hyperedge pairs, the probability of membership is the same, i.e.
\begin{equation*}
\Prob(u\in e)=p.
\end{equation*}

On the other hand, the CL procedure generates $\g^r$ with similar vertex degree and hyperedge size distribution as of $\g_1$.  The probability a vertex belongs to a hyperedge in $\g^r$ is proportional to the product of the desired vertex degree and hyperedge size, i.e.
\begin{equation*}
\Prob(u\in e)=\frac{d(u) d(e)}{c}.
\end{equation*}
To ensure this probability is always less than 1, one may further require the input sequences satisfy $\max_{i,j} d(u_i) d(e_j)\leq c$. Note that this procedure will in general produce a non-uniform hypergraph depending on distribution  of $d^1_e$. For a $k-$ uniform hypergraph, $d(e_i)=k,i=1,\cdots,m$. To sample a $k-$ uniform hypergraph with given node degree distribution $d^1_v$, we modify the CL process as follows:
\begin{itemize}
  \item [i.] Assign each node a probability $p_i$ as:
  \begin{equation*}
    p_i=\frac{d(v_i)}{c},\quad i=1,\cdots,\Nnodes.
  \end{equation*}
  \item [ii.] Select $k-$ distinct vertices with probabilities $p_{i_1},\cdots,p_{i_k}$. If the hypergraph does not already contain a hyperedge of the chosen $k$ vertices, then add the hyperedge to the hypergraph.
  \item [iii.] Repeat step ii) $m$ times.
\end{itemize}

\subsubsection{Mouse Neuron Endomicroscopy}
The mouse endomicroscopy dataset is an imaging video created under 10-minute periods of feeding, fasting and re-feeding using fluorescence across space and time in a mouse hypothalamus \cite{9119161,sweeney2021network,chen2021controllability}. Twenty neurons are recorded with individual levels of ``firing''. Similar to \cite{9119161}, we want to quantitatively differentiate the three phases. First, we compute the multi-correlation among every three neurons, which is defined by
\begin{equation}\label{eq:99}
    \rho = (1-\det{(\textbf{R}}))^{\frac{1}{2}},
\end{equation}
where, $\textbf{R}\in\mathbb{R}^{3\times 3}$ is the correlation matrix of three neuron activity levels \cite{wang2014measures}. When the multi-correlation $\rho$ is greater than a prescribed threshold, we build a hyperedge among the three neurons and assign it an hyperedge weight equal to $\rho$. We use a threshold of $0.93$ as used in \cite{9119161} for the purpose.

\begin{figure}[hbt!]
    \centering
    \includegraphics[trim=2cm 8cm 2cm 7.5cm,clip,scale=0.55]{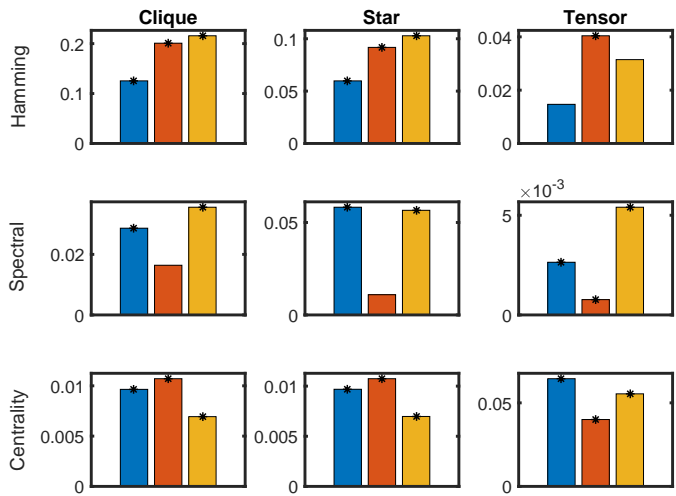}
    \caption{Comparison of different phases of mouse feeding activity using different HDMs: blue (fed-fast), red (fed-refed), and orange (refed-fast). The * on the bars implies that there is statistically significant difference between the hypergraphs in the corresponding two phases based on $p$-values from the permutation test.}
    \label{fig:mousebrain}
\end{figure}

Figure \ref{fig:mousebrain} shows the comparison of hypergraph corresponding to different phases using various HDMs. We find similar trends using both indirect and direct HDMs. The spectral and centrality HDMs between fed and refed phases have smaller values revealing more similarity at global and mesoscopic scales compared to corresponding HDM values between fed and fast, and refed and fast phases. On the other hand, Hamming HDM reveals that fed and fast phase are more similar at local scale compared to fed and refed phases. The * on the bars implies that there is statistically significant difference between the hypergraphs in the corresponding two phases based on $p$-values from the permutation test.

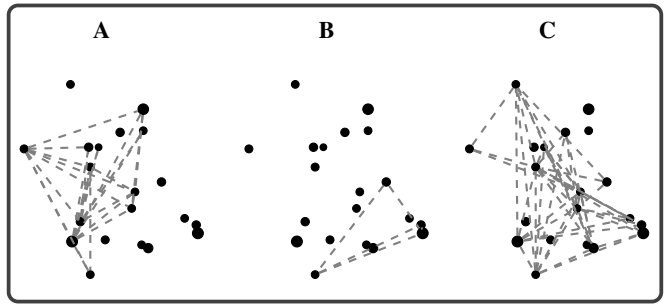
\begin{figure}[t]
    \centering
    \tcbox[colback=white,top=0pt,left=-1pt,right=1pt,bottom=5pt]{
    \begin{tikzpicture}[scale=0.022]
    \node[circle,inner sep=0.8] at  (60,150) {\small\textbf{A}};

    \node[circle,draw,fill=black,color=black,inner sep=4.12311/4] (v1) at  (110,36) {};
    \node[circle,draw,fill=black,color=black,inner sep=4.24264/4] (v2) at  (13,78) {};
    \node[circle,draw,fill=black,color=black,inner sep=4.47214/4] (v3) at  (96,58) {};
    \node[circle,draw,fill=black,color=black,inner sep=5.65685/4] (v4) at  (85,102) {};
    \node[circle,draw,fill=black,color=black,inner sep=5.83095/4] (v5) at  (42,22) {};
    \node[circle,draw,fill=black,color=black,inner sep=4.47214/4] (v6) at  (47,34) {};
    \node[circle,draw,fill=black,color=black,inner sep=4.24264/4] (v7) at  (53,67) {};
    \node[circle,draw,fill=black,color=black,inner sep=4.12311/4] (v8) at  (85,89) {};
    \node[circle,draw,fill=black,color=black,inner sep=4.12311/4] (v9) at  (78,42) {};
    \node[circle,draw,fill=black,color=black,inner sep=4.12311/4] (v10) at  (53,2) {};
    \node[circle,draw,fill=black,color=black,inner sep=4.24264/4] (v11) at  (62,23) {};
    \node[circle,draw,fill=black,color=black,inner sep=3.60555/4] (v12) at  (58,79) {};
    \node[circle,draw,fill=black,color=black,inner sep=4.47214/4] (v13) at  (52,79) {};
    \node[circle,draw,fill=black,color=black,inner sep=4.47214/4] (v14) at  (71,88) {};
    \node[circle,draw,fill=black,color=black,inner sep=4.12311/4] (v15) at  (80,52) {};
    \node[circle,draw,fill=black,color=black,inner sep=4.47214/4] (v16) at  (117,32) {};
    \node[circle,draw,fill=black,color=black,inner sep=4.12311/4] (v17) at  (41,117) {};
    \node[circle,draw,fill=black,color=black,inner sep=5/4] (v18) at  (88,18) {};
    \node[circle,draw,fill=black,color=black,inner sep=5.83095/4] (v19) at  (118,27) {};
    \node[circle,draw,fill=black,color=black,inner sep=4.12311/4] (v20) at  (84,20) {};

    \node[vertex,above of=v1,node distance=10pt,scale=0.7, white,opacity=0] (w1) {};
    \node[vertex,above of=v3,node distance=10pt,scale=0.7, white,opacity=0] (w3) {};
    \node[vertex,above of=v11,node distance=10pt,scale=0.7, white,opacity=0] (w11) {};
    \node[vertex,above of=v14,node distance=10pt,scale=0.7, white,opacity=0] (w14) {};
    \node[vertex,above of=v16,node distance=10pt,scale=0.7, white,opacity=0] (w16) {};
    \node[vertex,above of=v17,node distance=10pt,scale=0.7, white,opacity=0] (w17) {};
    \node[vertex,above of=v18,node distance=10pt,scale=0.7, white,opacity=0] (w18) {};
    \node[vertex,above of=v19,node distance=10pt,scale=0.7, white,opacity=0] (w19) {};
    \node[vertex,above of=v20,node distance=10pt,scale=0.7, white,opacity=0] (w20) {};
    \node[vertex,above of=v2,node distance=10pt,scale=0.7, white,opacity=0] (w2) {};
    \node[vertex,above of=v4,node distance=10pt,scale=0.7, white,opacity=0] (w4) {};
    \node[vertex,above of=v8,node distance=10pt,scale=0.7, white,opacity=0] (w8) {};


    \draw [dashed, thick,gray] (v2) -- (v4);
    \draw [dashed, thick,gray] (v2) -- (v5);
    \draw [dashed, thick,gray] (v2) -- (v6);
    \draw [dashed, thick,gray] (v2) -- (v7);
    \draw [dashed, thick,gray] (v2) -- (v9);
    \draw [dashed, thick,gray] (v2) -- (v10);
    \draw [dashed, thick,gray] (v2) -- (v13);
    \draw [dashed, thick,gray] (v2) -- (v15);
    \draw [dashed, thick,gray] (v4) -- (v5);
    \draw [dashed, thick,gray] (v4) -- (v9);
    \draw [dashed, thick,gray] (v4) -- (v15);
    \draw [dashed, thick,gray] (v5) -- (v6);
    \draw [dashed, thick,gray] (v5) -- (v7);
    \draw [dashed, thick,gray] (v5) -- (v8);
    \draw [dashed, thick,gray] (v5) -- (v9);
    \draw [dashed, thick,gray] (v5) -- (v10);
    \draw [dashed, thick,gray] (v5) -- (v12);
    \draw [dashed, thick,gray] (v5) -- (v13);
    \draw [dashed, thick,gray] (v5) -- (v15);
    \draw [dashed, thick,gray] (v6) -- (v7);
    \draw [dashed, thick,gray] (v6) -- (v8);
    \draw [dashed, thick,gray] (v6) -- (v12);
    \draw [dashed, thick,gray] (v6) -- (v13);
    \draw [dashed, thick,gray] (v6) -- (v15);
    \draw [dashed, thick,gray] (v7) -- (v10);
    \draw [dashed, thick,gray] (v7) -- (v15);
    \draw [dashed, thick,gray] (v9) -- (v15);

     \end{tikzpicture}
     \hspace{0.2cm}
     \begin{tikzpicture}[scale=0.022]
     \node[circle,inner sep=0.8] at  (60,150) {\small\textbf{B}};
    \node[circle,draw,fill=black,color=black,inner sep=4.12311/4] (v1) at  (110,36) {};
    \node[circle,draw,fill=black,color=black,inner sep=4.24264/4] (v2) at  (13,78) {};
    \node[circle,draw,fill=black,color=black,inner sep=4.47214/4] (v3) at  (96,58) {};
    \node[circle,draw,fill=black,color=black,inner sep=5.65685/4] (v4) at  (85,102) {};
    \node[circle,draw,fill=black,color=black,inner sep=5.83095/4] (v5) at  (42,22) {};
    \node[circle,draw,fill=black,color=black,inner sep=4.47214/4] (v6) at  (47,34) {};
    \node[circle,draw,fill=black,color=black,inner sep=4.24264/4] (v7) at  (53,67) {};
    \node[circle,draw,fill=black,color=black,inner sep=4.12311/4] (v8) at  (85,89) {};
    \node[circle,draw,fill=black,color=black,inner sep=4.12311/4] (v9) at  (78,42) {};
    \node[circle,draw,fill=black,color=black,inner sep=4.12311/4] (v10) at  (53,2) {};
    \node[circle,draw,fill=black,color=black,inner sep=4.24264/4] (v11) at  (62,23) {};
    \node[circle,draw,fill=black,color=black,inner sep=3.60555/4] (v12) at  (58,79) {};
    \node[circle,draw,fill=black,color=black,inner sep=4.47214/4] (v13) at  (52,79) {};
    \node[circle,draw,fill=black,color=black,inner sep=4.47214/4] (v14) at  (71,88) {};
    \node[circle,draw,fill=black,color=black,inner sep=4.12311/4] (v15) at  (80,52) {};
    \node[circle,draw,fill=black,color=black,inner sep=4.47214/4] (v16) at  (117,32) {};
    \node[circle,draw,fill=black,color=black,inner sep=4.12311/4] (v17) at  (41,117) {};
    \node[circle,draw,fill=black,color=black,inner sep=5/4] (v18) at  (88,18) {};
    \node[circle,draw,fill=black,color=black,inner sep=5.83095/4] (v19) at  (118,27) {};
    \node[circle,draw,fill=black,color=black,inner sep=4.12311/4] (v20) at  (84,20) {};

    \node[vertex,above of=v1,node distance=10pt,scale=0.7, white,opacity=0] (w1) {};
    \node[vertex,above of=v2,node distance=10pt,scale=0.7, white,opacity=0] (w2) {};
    \node[vertex,above of=v4,node distance=10pt,scale=0.7, white,opacity=0] (w4) {};
    \node[vertex,above of=v5,node distance=10pt,scale=0.7, white,opacity=0] (w5) {};
    \node[vertex,above of=v6,node distance=10pt,scale=0.7, white,opacity=0] (w6) {};
    \node[vertex,above of=v7,node distance=10pt,scale=0.7, white,opacity=0] (w7) {};
    \node[vertex,above of=v8,node distance=10pt,scale=0.7, white,opacity=0] (w8) {};
    \node[vertex,above of=v9,node distance=10pt,scale=0.7, white,opacity=0] (w9) {};
    \node[vertex,above of=v11,node distance=10pt,scale=0.7, white,opacity=0] (w11) {};
    \node[vertex,above of=v12,node distance=10pt,scale=0.7, white,opacity=0] (w12) {};
    \node[vertex,above of=v13,node distance=10pt,scale=0.7, white,opacity=0] (w13) {};
    \node[vertex,above of=v14,node distance=10pt,scale=0.7, white,opacity=0] (w14) {};
    \node[vertex,above of=v15,node distance=10pt,scale=0.7, white,opacity=0] (w15) {};
    \node[vertex,above of=v17,node distance=10pt,scale=0.7, white,opacity=0] (w17) {};
    \node[vertex,above of=v18,node distance=10pt,scale=0.7, white,opacity=0] (w18) {};
    \node[vertex,above of=v20,node distance=10pt,scale=0.7, white,opacity=0] (w20) {};

    \node[vertex,above of=v19,node distance=10pt,scale=0.7, white,opacity=0] (w19) {};
    \node[vertex,above of=v3,node distance=10pt,scale=0.7, white,opacity=0] (w3) {};


    \draw [dashed, thick,gray] (v3) -- (v10);
    \draw [dashed, thick,gray] (v3) -- (v19);
    \draw [dashed, thick,gray] (v10) -- (v16);
    \draw [dashed, thick,gray] (v10) -- (v19);
    \draw [dashed, thick,gray] (v16) -- (v19);

     \end{tikzpicture}
     \hspace{0.2cm}
     \begin{tikzpicture}[scale=0.022]
     \node[circle,inner sep=0.8] at  (60,150) {\small\textbf{C}};
    \node[circle,draw,fill=black,color=black,inner sep=4.12311/4] (v1) at  (110,36) {};
    \node[circle,draw,fill=black,color=black,inner sep=4.24264/4] (v2) at  (13,78) {};
    \node[circle,draw,fill=black,color=black,inner sep=4.47214/4] (v3) at  (96,58) {};
    \node[circle,draw,fill=black,color=black,inner sep=5.65685/4] (v4) at  (85,102) {};
    \node[circle,draw,fill=black,color=black,inner sep=5.83095/4] (v5) at  (42,22) {};
    \node[circle,draw,fill=black,color=black,inner sep=4.47214/4] (v6) at  (47,34) {};
    \node[circle,draw,fill=black,color=black,inner sep=4.24264/4] (v7) at  (53,67) {};
    \node[circle,draw,fill=black,color=black,inner sep=4.12311/4] (v8) at  (85,89) {};
    \node[circle,draw,fill=black,color=black,inner sep=4.12311/4] (v9) at  (78,42) {};
    \node[circle,draw,fill=black,color=black,inner sep=4.12311/4] (v10) at  (53,2) {};
    \node[circle,draw,fill=black,color=black,inner sep=4.24264/4] (v11) at  (62,23) {};
    \node[circle,draw,fill=black,color=black,inner sep=3.60555/4] (v12) at  (58,79) {};
    \node[circle,draw,fill=black,color=black,inner sep=4.47214/4] (v13) at  (52,79) {};
    \node[circle,draw,fill=black,color=black,inner sep=4.47214/4] (v14) at  (71,88) {};
    \node[circle,draw,fill=black,color=black,inner sep=4.12311/4] (v15) at  (80,52) {};
    \node[circle,draw,fill=black,color=black,inner sep=4.47214/4] (v16) at  (117,32) {};
    \node[circle,draw,fill=black,color=black,inner sep=4.12311/4] (v17) at  (41,117) {};
    \node[circle,draw,fill=black,color=black,inner sep=5/4] (v18) at  (88,18) {};
    \node[circle,draw,fill=black,color=black,inner sep=5.83095/4] (v19) at  (118,27) {};
    \node[circle,draw,fill=black,color=black,inner sep=4.12311/4] (v20) at  (84,20) {};

    \node[vertex,above of=v4,node distance=10pt,scale=0.7, white,opacity=0] (w4) {};
    \node[vertex,above of=v6,node distance=10pt,scale=0.7, white,opacity=0] (w6) {};
    \node[vertex,above of=v8,node distance=10pt,scale=0.7, white,opacity=0] (w8) {};
    \node[vertex,above of=v11,node distance=10pt,scale=0.7, white,opacity=0] (w11) {};
    \node[vertex,above of=v13,node distance=10pt,scale=0.7, white,opacity=0] (w13) {};
    \node[vertex,above of=v20,node distance=10pt,scale=0.7, white,opacity=0] (w20) {};
    \node[vertex,above of=v1,node distance=10pt,scale=0.7, white,opacity=0] (w1) {};
    \node[vertex,above of=v5,node distance=10pt,scale=0.7, white,opacity=0] (w5) {};
    \node[vertex,above of=v10,node distance=10pt,scale=0.7, white,opacity=0] (w10) {};


    \draw [dashed, thick,gray] (v1) -- (v5);
    \draw [dashed, thick,gray] (v1) -- (v7);
    \draw [dashed, thick,gray] (v1) -- (v9);
    \draw [dashed, thick,gray] (v1) -- (v15);
    \draw [dashed, thick,gray] (v2) -- (v7);
    \draw [dashed, thick,gray] (v2) -- (v15);
    \draw [dashed, thick,gray] (v2) -- (v17);
    \draw [dashed, thick,gray] (v3) -- (v7);
    \draw [dashed, thick,gray] (v3) -- (v9);
    \draw [dashed, thick,gray] (v3) -- (v17);
    \draw [dashed, thick,gray] (v5) -- (v7);
    \draw [dashed, thick,gray] (v5) -- (v9);
    \draw [dashed, thick,gray] (v5) -- (v10);
    \draw [dashed, thick,gray] (v5) -- (v15);
    \draw [dashed, thick,gray] (v5) -- (v16);
    \draw [dashed, thick,gray] (v5) -- (v17);
    \draw [dashed, thick,gray] (v7) -- (v9);
    \draw [dashed, thick,gray] (v7) -- (v10);
    \draw [dashed, thick,gray] (v7) -- (v12);
    \draw [dashed, thick,gray] (v7) -- (v14);
    \draw [dashed, thick,gray] (v7) -- (v15);
    \draw [dashed, thick,gray] (v7) -- (v16);
    \draw [dashed, thick,gray] (v7) -- (v17);
    \draw [dashed, thick,gray] (v7) -- (v18);
    \draw [dashed, thick,gray] (v7) -- (v19);
    \draw [dashed, thick,gray] (v9) -- (v10);
    \draw [dashed, thick,gray] (v9) -- (v12);
    \draw [dashed, thick,gray] (v9) -- (v14);
    \draw [dashed, thick,gray] (v9) -- (v15);
    \draw [dashed, thick,gray] (v9) -- (v16);
    \draw [dashed, thick,gray] (v9) -- (v17);
    \draw [dashed, thick,gray] (v9) -- (v18);
    \draw [dashed, thick,gray] (v9) -- (v19);
    \draw [dashed, thick,gray] (v10) -- (v12);
    \draw [dashed, thick,gray] (v10) -- (v14);
    \draw [dashed, thick,gray] (v10) -- (v15);
    \draw [dashed, thick,gray] (v10) -- (v17);
    \draw [dashed, thick,gray] (v10) -- (v18);
    \draw [dashed, thick,gray] (v10) -- (v19);
    \draw [dashed, thick,gray] (v12) -- (v15);
    \draw [dashed, thick,gray] (v12) -- (v15);
    \draw [dashed, thick,gray] (v12) -- (v18);
    \draw [dashed, thick,gray] (v14) -- (v18);
    \draw [dashed, thick,gray] (v15) -- (v16);
    \draw [dashed, thick,gray] (v15) -- (v17);
    \draw [dashed, thick,gray] (v15) -- (v18);
     \draw [dashed, thick,gray] (v17) -- (v18);
     \draw [dashed, thick,gray] (v18) -- (v19);
    \end{tikzpicture}
    }

\hspace{1cm}
    \caption{Mouse neuron endomicroscopy features. (A), (B) and (C) Neuronal activity networks of the three phases - fed, fast and re-fed, which depicts the spatial location and size of individual neurons. Each 2-simplex (i.e., a triangle) represents a hyperedge. The cutoff threshold is 0.93 for the hypergraph model.}
    \label{fig:mouse}
\end{figure}

\subsubsection{Genomic Dataset}
We next apply HDM framework to compare genomic structure of two cell types. Genomic DNA must be folded
to fit inside a nucleus, but must remain accessible for gene transcription, replication and repair \cite{rajapakse2011emerging,chen2015functional}.
Consequently, higher-order chromatin structure arises from such combinatorial physical interactions of
many genomic loci. Recently, authors in \cite{poreC} proposed to represent such higher-order
chromatin structure by a hypergraph, where the different loci in the genome are the vertices, and each
multi-way contact between a set of loci represent a hyperedge. Furthermore, they used Pore-C \cite{howorka2020reading}, a recent method developed by Oxford Nanopore Technologies to measure these multi-way contacts directly and construct
the hypergraph experimentally. Note that, while Pore-C gives contact information at the finest level of base-pair position in the genome, it is often convenient to aggregate this information at a coarser resolution by aggregating linear continuous segments in the genome, see \cite{poreC} for details.

\begin{figure*}[hbt!]
    \centering
    \includegraphics[trim=1.5cm 10.5cm 0cm 0cm,clip,scale=0.58]{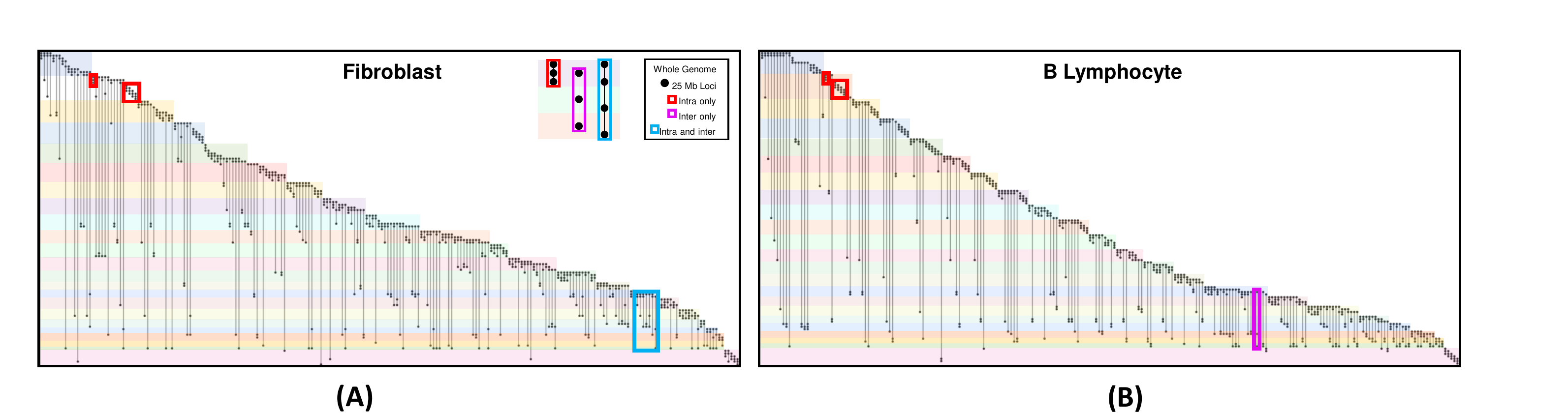}
    \caption{(A) Incidence matrix visualization of the top 10 most common multi-way contacts per chromosome, for fibroblasts. Matrices are constructed at 25 Mb resolution. Highlighted boxes indicate example intra-chromosomal contacts (red), inter-chromosomal contacts (magenta), and combinations of intra- and inter-chromosomal contacts (blue). Examples for each type of contact are shown in the top right corner.  Genomic loci that do not participate in the top 10 most common multi-way contacts for fibroblasts or B lymphocytes were removed from the incidence plots. (B) Similar plot for B lymphocytes. }
    \label{fig:genome_inter}
\end{figure*}

Figure \ref{fig:genome_inter} shows the visualization of the incidence matrix of the hypergraph derived for human fibroblasts (FB) and B lymphocytes (GM) cell lines at $25$Mb resolution after noise reduction. The entire genome at this resolution consists of $\Nnodes=3,102$ vertices for both cells, with number of hyperedges $m=836,571$ for FM and $m=1,028,694$ for GM. The maximum hyperedge set cardinality is $40$ for FB and $90$ for GM. To compare chromosomes individually, we also construct separate hypegraphs for each chromosome comprising of intra-chromosomal contacts only. Chromosome $1$ has maximum number of vertices $\Nnodes=249$ and chromosome $22$ has smallest number of vertices $\Nnodes=51$ at the chosen $25$Mb scale. The number of hyperedges differ by chromosomes taking values in range $[1,000 \quad 35,000]$ for FM, and in the range $[6,000 \quad 75,000]$  for GM, respectively. Moreover maximum hyperedge set cardinality also differs between corresponding chromosomes in the GM and FB. As a result we cannot use the tensor based direct HDMs, and restrict to indirect HDM for comparison.

\begin{figure}
  \begin{center}
  \includegraphics[trim=2.5cm 7.5cm 2.5cm 7.5cm,clip,scale=0.55]{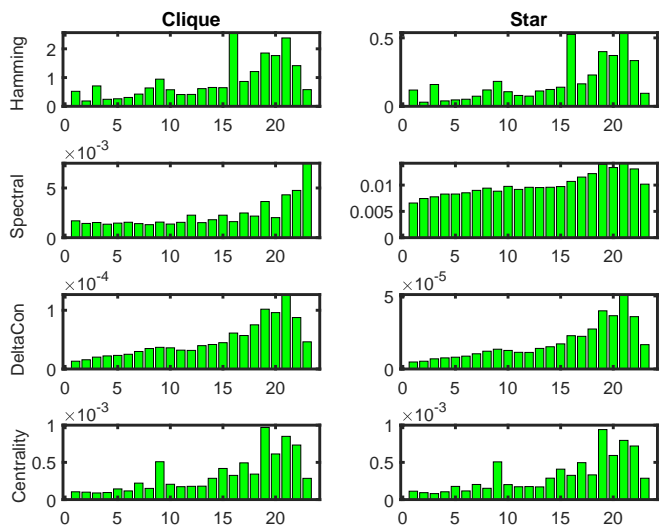}
  \caption{Comparison of different chromosomes in FB and GM using indirect HDMs.}\label{fig:chrcmp}
  \end{center}
\end{figure}

In Figure \ref{fig:chrcmp} we show chromosome level comparison using indirect HDM based on the clique and star expansion. We find that the trends between two expansions are similar for Hamming, deltaCon, and centrality HDMs, while they differ for spectral HDM. Furthermore, we can see that at local scale chromosome $16$ differs most between FB and GM, while chromosome $19$ and $21$ differ the most at the mesoscale. The clique-spectral HMD reveals that chromosome $23$ differs most at global scale, while star-spectral HMD indicates chromosomes $19,20,21,22$ are the most different.

\begin{table}
  \centering
  \begin{tabular}{|c|c|c|}
  \hline
          & Clique  & Star  \\\hline
  Hamming & $2.3\times 10^{-2} (*)$ & $2\times 10^{-3} (*)$ \\\hline
  Spectral & $6.2\times 10^{-4} (*)$  & $1.6 \times 10^{-3} (*)$ \\\hline
  deltaCon & $4.7\times 10^{-7} (*)$ & $1.6\times 10^{-7} (*)$  \\\hline
  Centrality & $2.5\times 10^{-6} (*)$ & $2.9\times 10^{-6} (*)$ \\\hline
  \end{tabular}
  \caption{Indirect HDMs values between full genome of the FB and GM. The * in bracket implies that difference between FB and GM is statistically significant based on $p-$ values from the permutation test.}\label{tableDM}
\end{table}

Table \ref{tableDM} shows the values of different indirect HDM between FB and GM for the entire genome. The $p$-values which suggest that FB and GM are dissimilar at all scales based on clique expansion. On the other hand, using star expansion, FB and GM are different at local and global scales, and look similar at mesoscopic scale.

\subsection{Discussion}\label{sec:discussion}
We first discuss pros/cons of indirect and direct HDMs. While indirect HDMs allow one to leverage large variety of GDMs for hypergraph comparison, the hypergraph conversion into clique/star representation is lossy and thus may result in inability to discern certain aspects of structural differences or similarities between two hypergraphs. Direct HDMs being based on tensor representation should not suffer from such limitation. However, tensor computations (e.g., tensor eigenvalue/singular values) can be challenging for hypergraphs with large number of vertices and/or with high maximum hyperedge cardinality. Moreover, direct HDMs can only be applied to cases where the underlying hypergraph has same number of vertices and same maximum hyperedge cardinality. Indirect HDMs however are computationally less demanding, and can be employed even if hypergraphs have different maximum hyperedge cardinality. Moreover, by restricting to GDMs which are applicable for comparing graphs with different number of vertices or unknown node correspondence, indirect HDMs can also be applied for comparing hypergraphs with different number of vertices and/or unknown node correspondence. In terms of performance of indirect and direct HDMs to assess structural differences or similarities between two hypergraphs, the numerical studies show that both approaches could be effective depending on the application.

We are currently exploring the application of line expansion \cite{yang2020hypergraph} which has been recently proposed as an alternative approach to transforming hypergraph into a graph. Compared to clique or star expansion, line expansion does not result in any information loss during the transformation, thus, potentially providing more effective means for developing indirect HDMs. Addressing the computational challenges associated with tensor based HDM will be important to address to scale the approach to larger problems. In addition approaches alternative to using tensor based representation, such as higher order random walks based hypergraph analysis \cite{aksoy2020hypernetwork} provide another potential avenue for developing new HDMs. While notions of graph kernels and graph embedding have been extended to hypergraphs \cite{payne2019deep,feng2019hypergraph,lugo2017classification,bai2021hypergraph}, further investigation is warranted for their application in hypergraph comparison.  Applications of the proposed HDMs in other domains e.g. cyber security and social networks is another direction of future research.

\section{Conclusion}\label{sec:conc}
In this paper we presented two approaches for hypergraph comparison. The first approach transforms the hypergraph into a graph representation, and then uses standard graph dissimilarity measures. The second approach uses tensors to represent hypergraphs and  then invokes various tensor algebraic notions to develop hypergraph dissimilarity measures. Within each approach we presented a collection of measures which  assess hypergraph dissimilarity at different scales. We evaluated these measures on synthetic hypergraphs, and real world biological datasets with promising results. Finally, we discussed various pros/cons in using the two approaches, and outlined some avenues of future research.

\section*{Acknowledgments}
This material is based upon work supported by the Air Force Office of Scientific Research under award number FA9550-18-1-0028. Any opinions, finding, and conclusions or recommendations expressed in this material are those of the author(s) and do not necessarily reflect the views of the United States Air Force.

\ifCLASSOPTIONcaptionsoff
  \newpage
\fi



\bibliographystyle{IEEEtran}
\bibliography{tensorref}
\end{document}